\documentclass[10pt,twocolumn,letterpaper]{article}

\usepackage[]{cvpr}      
\usepackage{amssymb, amsmath, multirow, enumitem, colortbl, fancyhdr}
\usepackage{bbding}
\usepackage{caption}
\usepackage{subcaption}
\usepackage{wrapfig} 
\usepackage[ruled,vlined]{algorithm2e}
\usepackage{algorithmic}
\usepackage{graphicx}
\usepackage{svg}
\usepackage{microtype}
\usepackage[accsupp]{axessibility}
\usepackage[none]{hyphenat}
\definecolor{tabbestgreen}{HTML}{4CAF50}      
\definecolor{tabsecondgreen}{HTML}{81C784}   
\definecolor{bestgreen}{HTML}{4CAF50}      
\definecolor{secondgreen}{HTML}{81C784}   
\definecolor{secondblue}{HTML}{90CAF9}
\definecolor{figblue}{rgb}{0, 0.6902, 0.9412}
\definecolor{figorange}{rgb}{0.9294, 0.4902, 0.1921}
\newcommand{\best}{\cellcolor{bestgreen}}
\newcommand{\sbest}{\cellcolor{secondblue}}
\usepackage[normalem]{ulem}

\usepackage[dvipsnames]{xcolor}

\definecolor{tableyellow}{rgb}{1, 1, 0.7}
\definecolor{tableorange}{rgb}{1, 0.85, 0.7}
\definecolor{tablered}{rgb}{1, 0.7, 0.7}

\definecolor{tabfirst}{rgb}{1, 0.7, 0.7}
\definecolor{tabsecond}{rgb}{1, 0.85, 0.7}
\definecolor{tabthird}{rgb}{1, 1, 0.7}

\definecolor{cvprblue}{rgb}{0.21,0.49,0.74}
\usepackage[pagebackref,breaklinks,colorlinks,allcolors=cvprblue]{hyperref}

\def\paperID{} 
\def\confName{CVPR}
\def\confYear{2026}

\title{ReLaGS: Relational Language Gaussian Splatting}

\author{
Yaxu Xie$^{1,2}$ $^*$  \quad
Abdalla Arafa$^{1,2}$  $^*$  \quad
Alireza Javanmardi$^{1}$ \quad
Christen Millerdurai$^{1}$ \\
Jia Cheng Hu$^{3}$ \quad
Shaoxiang Wang$^{1,2}$ \quad
Alain Pagani$^{1}$ \quad
Didier Stricker$^{1,2}$ \\
$^{1}$German Research Center for Artificial Intelligence (DFKI) \\
$^{2}$RPTU University Kaiserslautern-Landau \quad 
$^{3}$University of Modena and Reggio Emilia 
}

\begin{document}

\twocolumn[{
    \renewcommand\twocolumn[1][]{#1}%
    \maketitle
    \vspace{-3.0em}
    \begin{center}
        \includegraphics[width=0.8\textwidth]{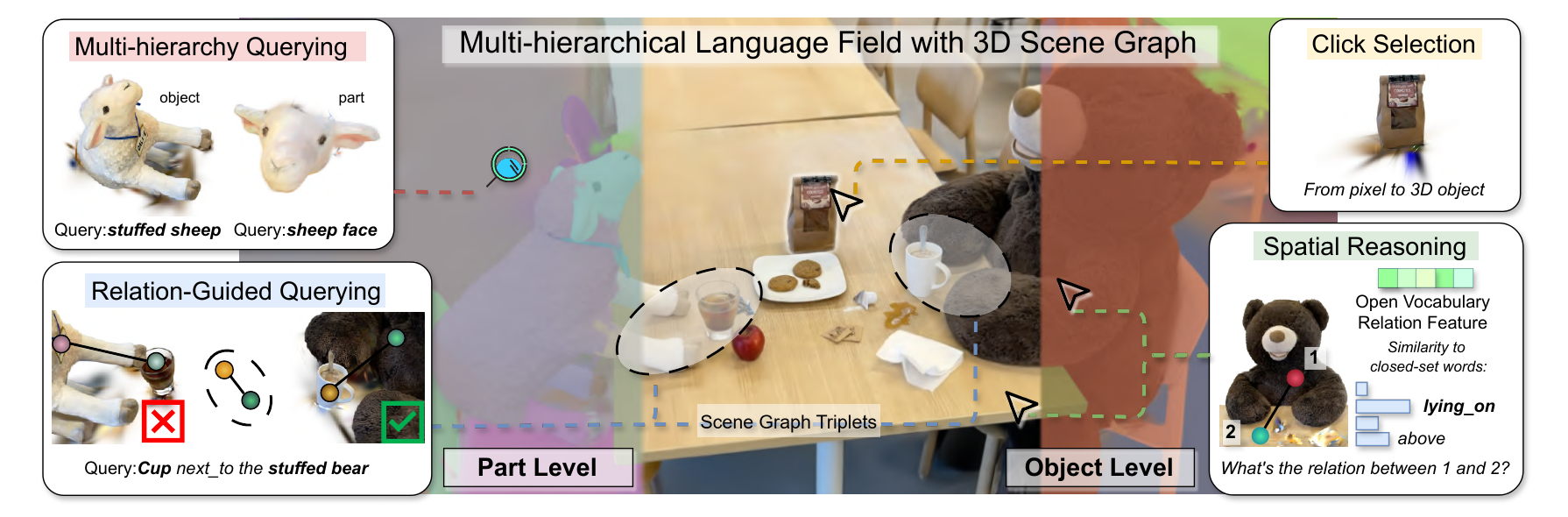}
        \captionof{figure}{
        \textbf{Relational Language Gaussian Splatting}.
        We build a platform with multi-hierarchical language Gaussian field and open-vocabulary 3D scene graph, to support various tasks such as object selection via click, open vocabulary 3D object segmentation across semantic granularity, spatial relationship reasoning between objects and querying object with relation-guidance. 
        }
        \label{fig:teaser} \vspace{-2mm}
    \end{center}
}]
\def\thefootnote{*}\footnotetext{These authors contributed equally to this work}
\begin{abstract}

Achieving unified 3D perception and reasoning across tasks such as segmentation, retrieval, and relation understanding remains challenging, as existing methods are either object-centric or rely on costly training for inter-object reasoning. 
We present a novel framework that constructs a hierarchical language-distilled Gaussian scene and its 3D semantic scene graph without scene-specific training. 
A Gaussian pruning mechanism refines scene geometry, while a robust multi-view language alignment strategy aggregates noisy 2D features into accurate 3D object embeddings. 
On top of this hierarchy, we build an open-vocabulary 3D scene graph with Vision Language-derived annotations and Graph Neural Network-based relational reasoning. 
Our approach enables efficient and scalable open-vocabulary 3D reasoning by jointly modeling hierarchical semantics and inter/intra-object relationships, validated across tasks including open-vocabulary segmentation, scene graph generation, and relation-guided retrieval.
Project page: \url{https://dfki-av.github.io/ReLaGS/}

\end{abstract}
\vspace{-5mm}    
\section{Introduction}
\label{sec:intro}

Radiance-field representations such as NeRF~\cite{mildenhall2021nerf} and Gaussian Splatting~\cite{kerbl20233d} have become core technologies for high-fidelity 3D reconstruction in VR/AR, robotics, and digital twins.
Despite their impressive geometric and photometric accuracy, these representations lack meaningful scene semantics and therefore cannot support high-level reasoning.
Recent \emph{language field distillation} methods~\cite{kerr2023lerf, engelmann2024opennerf, zhou2024feature, wu2024opengaussian, jun2025dr} address this gap by injecting vision–language priors into 3D radiance fields, enabling them to encode geometry, colors, and language embeddings distilled from 2D foundation models.
This transforms purely geometric radiance fields into open-vocabulary 3D feature fields that can be queried directly in natural language, supporting tasks such as open-vocabulary segmentation~\cite{qu2024goi, wu2024opengaussian}, language-guided navigation~\cite{chen2025splat, lei2025gaussnav}, and scene editing~\cite{qiu2024language, wang2026inpaint360gs}.
In such fields, users can issue queries such as \textit{``find the wooden chair''} or \textit{``highlight all green plants''} to retrieve semantically matched regions.
However, despite these capabilities, existing methods remain fundamentally limited in semantic expressiveness. They fail in queries involving spatial relations or object parts---such as
\textit{``select the cup next to the laptop''} or \textit{``highlight the keyboard of the laptop''}.
These systems often segment the entire laptop instead of its keyboard or ignore spatial configurations altogether, as they operate with a single semantic granularity and without relational context.
This reveals two key limitations of existing works: they remain \textbf{flat}, with single-level semantic abstraction  and \textbf{isolated}, lacking inter-entity relationships.
They describe what objects exist but not how they relate, because their features remain confined within individual entities.
Furthermore, these approaches lack a hierarchical organization of semantic abstraction, missing both finer part-level cues and higher-level contextual relationships essential for comprehensive and structured 3D scene understanding.

To overcome these limitations, recent efforts aim to embed relational context into radiance field~\cite{koch2025relationfield, wang2025gaussiangraph}, bridging toward 3D scene graph formulations.
RelationField~\cite{koch2025relationfield} models relationships as ray pairs connecting object instances, learning spatial and functional dependencies between entities via multimodal language priors, but its volumetric rendering pipeline remains computationally heavy and memory intensive.
SplatTalk~\cite{thai2025splattalk} converts language-distilled Gaussians into semantic tokens for Large Language Model (LLM) reasoning, but tokenization and LoRA~\cite{hu2022lora} fine-tuning remain costly and slow.

This puts us at a crossroads: we seek both multi-level scene representations capturing full \textit{part--object--scene} hierarchy and relational 3D scene graphs that explain how entities interact. 
Yet existing methods achieve only one of these goals or rely on computationally expensive optimization.
We therefore pursue both goals in a unified, optimization-free framework that organizes 3D scenes into coherent semantic hierarchies while also capturing their relational structure. To this end, we introduce a training-free pipeline that first constructs a language-grounded hierarchical Gaussian scene representation and then builds an explicit open-vocabulary 3D scene graph on top of it. 
Rather than assigning language and relation features to every Gaussian, we observe that semantics naturally emerge at coarser spatial granularity, while appearance details remain at the Gaussian level.
Following this intuition, we cluster Gaussians in a bottom-up manner into sub-part, part, and object-level groups, regulated by multi-level masks.
While segmentation-based methods~\cite{wu2024opengaussian, dai2025training} achieve strong geometric grouping, they weakly handle language registration under inconsistent SAM masks and noisy CLIP~\cite{radford2021learning} features. 
We address this challenge by introducing \emph{Maximum Weight Pruning} and \emph{Robust Outlier-Aware Feature Aggregation}, which jointly refine geometry and filter out unreliable language features, yielding consistent per-object language embeddings across multiple views and hierarchy levels.

On top of this hierarchy, we explicitly construct an open-vocabulary 3D scene graph that links entities through relational edges summarizing spatial and relationships.
Our design naturally extends to intra-object graphs when finer-grained relational understanding is required.
We propose two complementary variants for scene graph construction. 
The first lifts per-frame relational annotations obtained from LLM with Set-of-Mark (SoM) prompting~\cite{yang2023set} into 3D, providing high-quality but sparse relation edges. 
The second employs a pretrained graph transformer to infer relations directly from geometric and language features
without additional costs and ensuring scalability, as we mainly evaluated in the experiments.
Compared to RelationField, which requires hours of training and renders below 10~fps, our pipeline constructs a complete scene graph in under \textbf{15~minutes} and renders at over 200~fps.
In summary, our framework unifies hierarchical scene construction and relational reasoning within 3D Gaussian fields, providing an explicit, scalable, and training-free foundation for structured 3D understanding. 
Our main contributions are as follows:
\begin{itemize}
    \item We propose the first unified language distillation framework allowing for both hierarchical and relational reasoning on one single Gaussian field.
    \item We introduce Maximum Weight Gaussian pruning and robust outlier-aware feature aggregation methods, largely improving the geometric and language registration accuracy of the hierarchical Gaussian scene reconstruction pipeline.
    \item We provide two solutions to build explicit open-vocabulary 3D scene graphs: lifting 2D LLM-based annotations into 3D, and predicting relationships with a pretrained lightweight graph neural network.
    \item We demonstrate that our design achieves structured, open-vocabulary 3D scene understanding without scene-specific training, efficiently combining geometry, language, and relationships within a unified scene.
\end{itemize}

\section{Related Work and Motivation}
\label{sec:related_work}
\begin{figure*}[t]
 \centering
 \vspace{-8mm}
    \includegraphics[width=0.95\textwidth]{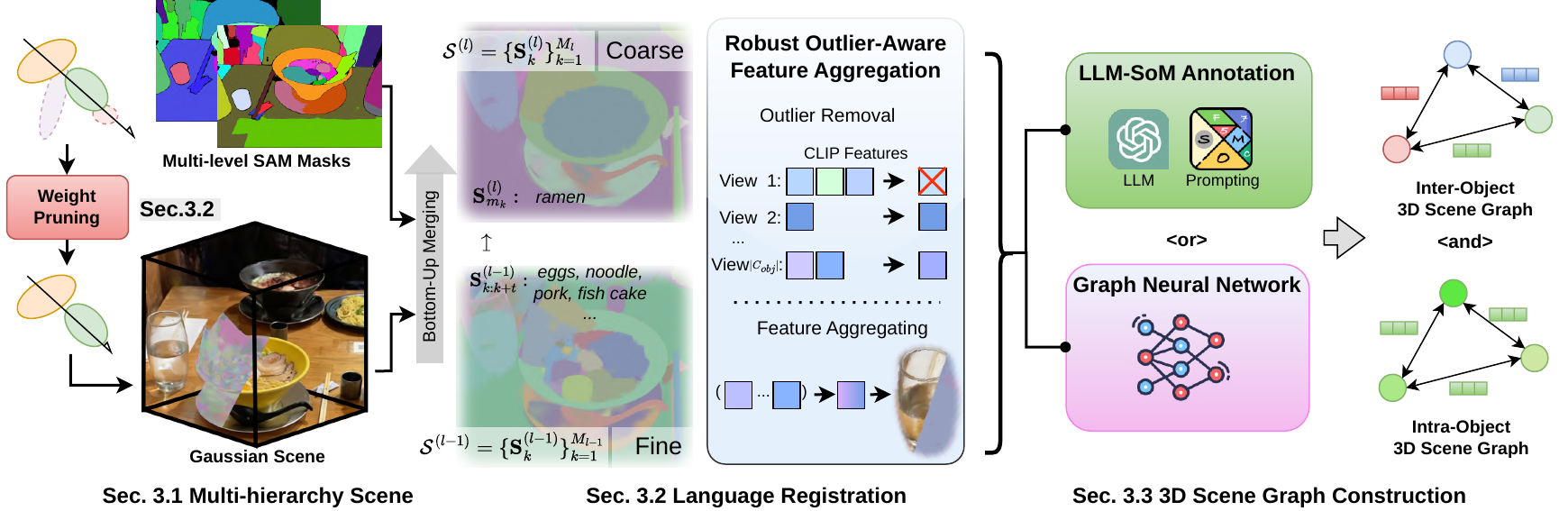}
        \caption{
        \textbf{ReLaGS Overview}.
        Given a reconstructed Gaussian scene, redundant primitives are first pruned to improve geometric accuracy. 
        Heuristic clustering under multi-level SAM supervision then forms a hierarchical scene structure, where each cluster is assigned a CLIP-based language feature with outlier rejection. 
        Finally, open-vocabulary inter- and intra-object scene graphs are obtained either by lifting LLM-derived relations for semantic diversity or by using a pretrained graph network for efficient offline inference.
        }
        \label{fig:overview}\vspace{-6mm}
\end{figure*}

\noindent\textbf{Language Field Distillation.}
Existing approaches for embedding language semantics into 3D Gaussian or radiance fields can be broadly categorized as \textit{training-based} and \textit{training-free}.
Training-based methods~\cite{kerr2023lerf,qin2024langsplat,shi2024language,zhou2024feature} incorporate vision–language supervision directly into the rendering and optimization loop of the radiance field.
Although effective, per-scene training is inefficient in time and resource, and remains highly sensitive to view-inconsistent 2D features, making them impractical for large-scale deployment.
To improve scalability, training-free and heuristic forward distillation methods~\cite{cheng2024occamslgssimpleapproach,jun2025dr,wang2025visibility,xiong2025splat} lift 2D feature maps into 3D Gaussians through weighted or closed-form aggregation without gradient updates.
Occam’s LGS~\cite{cheng2024occamslgssimpleapproach} formulates this as a MAP estimation problem solvable in closed form. Dr.Splat~\cite{jun2025dr} adopts a similar scheme with top-$k$ truncation for efficiency, and VALA~\cite{wang2025visibility} introduces a visibility-aware gating mechanism to suppress occluded contributions.
Splat Feature Solver~\cite{xiong2025splat} further interprets feature lifting as a sparse linear inverse problem and proves that weighted aggregation methods~\cite{jun2025dr,marrie2025ludvig,lee2025cf3} are bounded approximations to its least-squares solution.
THGS~\cite{dai2025training} stands out among these by extending the \textit{grouping-then-registration} strategy~\cite{wu2024opengaussian,liang2024supergseg} into a training-free paradigm.
It hierarchically merges Gaussians from superpoints to parts and objects, guided by multi-level SAM masks, forming a nested and interpretable structure that serves as the foundation for our scene construction.\\[1pt]
\noindent\textbf{3D Scene Graph}~\cite{armeni20193d, wald2020learning} is a structured graph representation that links objects, their spatial relationships, and semantic attributes in 3D scenes.
Although many 3D scene graph generation works~\cite{wald2020learning, wu2021scenegraphfusion, wu2023incremental, wang2023vl, lv2024sgformer} are limited by closed-set semantic categories and highly rely on geometric reasoning in pre-segmented point cloud, recent advances have moved toward open-vocabulary 3D scene graph generation~\cite{koch2024open3dsg, wu2025universal}, which integrates vision language models to associate 3D entities with rich semantics aligned to the language. 
ConceptGraphs~\cite{gu2024conceptgraphs}, GaussianGraph~\cite{wang2025gaussiangraph}, and RelationField~\cite{koch2025relationfield} all aim to build open-vocabulary 3D scene graphs by combining vision–language features with geometry.
ConceptGraphs depends on costly LLM inference and outputs text-based graphs,
GaussianGraph requires scene-specific training and encodes relations implicitly through geometric heuristics,
and RelationField learns relation fields via per-scene optimization, resulting in high computational cost and limited explicitness.

\noindent\textbf{Our Motivation} is to \textbf{\textit{efficiently}} construct an \textbf{\textit{explicit}} open-vocabulary 3D scene graph that connects the \textbf{\textit{hierarchical}} Gaussian language field for structured \textbf{\textit{causal reasoning}} in 3D. 
To ensure efficiency and explicitness, we adopt a training-free forward distillation framework~\cite{dai2025training} and employ a lightweight pretrained graph network to predict open-vocabulary relation embeddings from object language features and spatial cues. 
Alternatively, object-ID maps rendered from hierarchical scenes can be paired with SoM-prompted multimodal LLMs to annotate 2D relations, which are then lifted into 3D. 
Thus, inter-object relationships are represented explicitly as \textit{object–predicate–subject} triplets, where both nodes and edges are discrete, language-grounded entities within the Gaussian scene.
Because scene graph construction depends on accurate object segmentation and language registration, we further improve modules to ensure reliable relationship prediction. 
The hierarchical organization dictates how relationships are modeled across abstraction levels: inter-object graphs capture global spatial and functional relations, while intra-object graphs describe fine-grained part structures within each entity. 
Relations between parts of different objects (e.g., a \textit{door handle} and a \textit{table leg}) are generally meaningless and are instead summarized by their parent objects (e.g., \textit{table next to door}), ensuring coherent and interpretable scene graphs. 
This hierarchical separation not only preserves clarity but also supports \textbf{causal reasoning} across levels—linking local part composition to global spatial context—thereby enabling structured understanding of how entities and their interactions collectively form a 3D scene.

\section{Approach}
\label{sec:approach}
Our method, \textbf{ReLaGS}, unifies hierarchical scene representation and relational reasoning within a single framework without scene-specific training.
As illustrated in Fig.~\ref{fig:overview}, given a reconstructed 3D Gaussian field, ReLaGS first constructs a multi-hierarchy representation that organizes the scene into nested levels of semantic abstraction, from fine-grained parts to complete objects. 
This hierarchical structure provides a compact and interpretable foundation for reasoning, language registration, and relationship prediction. 
We then explicitly build 3D scene graphs whose nodes correspond to hierarchical entities and whose edges encode semantic relations, either lifted from large language models or predicted by a pretrained graph neural network. 
We next formalize each stage of this pipeline, beginning with the hierarchical Gaussian representation.

\subsection{Multi-Hierarchy Gaussian Representation}
\label{sec:approach:scene}

Let $\mathcal{G}=\{G_i\}_{i=1}^{N}$ denote the set of 3D Gaussian primitives, each encoding spatial and radiance attributes.
We define a hierarchical representation with $L$ abstraction levels:
\begin{equation}
    \mathcal{S}^{(1)}, \dots, \mathcal{S}^{(L)}, \quad 
    \mathcal{S}^{(l)}=\{\mathbf{S}^{(l)}_k\}_{k=1}^{M_l}, \quad k \in \{1,\dots,M_{l}\}
\end{equation}
where $\mathcal{S}^{(l)}$ is the set of clusters at level $l$ and $M_l$ is the number of clusters at that level.
Each Gaussian $G_i$ is assigned to exactly one cluster at each level, and the clusters form a nested hierarchy:
\begin{equation}
    \forall\, \mathbf{S}^{(l-1)}_{k} \in \mathcal{S}^{(l-1)}, \quad 
\mathbf{S}^{(l-1)}_{k} \subseteq \mathbf{S}^{(l)}_{m_k}, \quad 
\mathbf{S}^{(l)}_{m_k} \in \mathcal{S}^{(l)}.
\end{equation}
Each cluster $\mathbf{S}^{(l)}_k$ represents a semantic entity at level $l$.
For instance in Fig.~\ref{fig:overview}\, $\mathbf{S}^{(3)}_{m_k}$ correspond to a complete object (e.g., a \textit{ramen}), while lower levels such as $\mathbf{S}^{(2)}_k$ capture finer-grained parts (e.g., the \textit{eggs} as its ingredient).
Each cluster carries an embedding $\mathbf{f}^{(l)}_k \in \mathbb{R}^d$, which encodes its language-aligned semantics.
This induces a semantic hierarchy,
$
    \mathcal{G} \;\rightarrow\; \mathcal{S}^{(1)} \;\rightarrow\; \cdots \;\rightarrow\; \mathcal{S}^{(L)}
$, where higher levels represent coarse semantic granularity. 

\noindent \textbf{Rasterization and 2D-3D Tracing.}
To support tasks such as multi-hierarchy instance segmentation via point prompts on rendered images and lifting 2D scene graph annotations into 3D (see Sec.~\ref{sec:app:scene_graph}), we establish a consistent pixel-to-Gaussian correspondence through rasterization. 
The color $\mathbf{C}$ of each pixel $(u,v)$ is obtained by volumetric $\alpha$-blending along its viewing ray:
\begin{equation}
    \mathbf{C}_{(u,v)} = \sum_{i=1}^{N} \mathbf{c}_i \, \alpha_i \, T_i, 
    \quad T_i = \prod_{j<i} (1-\alpha_j),
\end{equation}
where $\mathbf{c}_i$ and $\alpha_i$ denote the color and opacity of Gaussian $G_i$, and $T_i$ represents the accumulated transmittance of all preceding Gaussians along the ray.
We compute the contribution weight $w_i = \alpha_i T_i$ for each Gaussian and define the dominant Gaussian for pixel $(u,v)$ as:
\begin{equation}
    G^{\text{*}}_{(u,v)} = \arg\max_{i} \, w_i.
    \label{eq:max_contrib}
\end{equation}
This dominant Gaussian provides a reliable 2D-3D correspondence, which naturally extends to map a pixel to cluster $G^{\text{*}}_{(u,v)} \in \mathbf{S}^{(l)}_{k}$.

\subsection{Improving Construction and Language Lifting of Hierarchical Scene}

To construct the multi-hierarchy scene representation described in Sec.~\ref{sec:approach:scene}, we adopt THGS~\cite{dai2025training}, which organizes the Gaussian field without gradient-based optimization.
The generated field $\mathcal{G}$ is partitioned into geometrically coherent superpoints via Cut Pursuit~\cite{landrieu2017cut} and progressively merged into higher-level clusters $\mathcal{S}^{(1)}, \mathcal{S}^{(2)}, \dots, \mathcal{S}^{(L)}$ by balancing intra-cluster compactness and inter-cluster boundary saliency, guided by SAM mask priors~\cite{kirillov2023segment}.
Further details can be found in THGS~\cite{dai2025training}---Secs.~3.2–3.4.

Since accurate segmentation and language feature of objects are both critical for scene graph prediction and reconstruction,
we augment THGS with two novel complementary steps that enhance the multi-hierarchy scene representation by refining both geometric and semantic quality,
yielding more accurate object–part boundaries and improved language-aligned object representations. First, Maximum Weight Pruning~(MWP) eliminates Gaussians with negligible visual impact across all training views, ensuring that only geometrically relevant components are retained. Second, Robust Outlier-Aware Feature Aggregation~(ROFA) enhances the reliability of object-level semantics by filtering inconsistent CLIP features before aggregation. Together, these steps yield a compact yet semantically consistent representation that serves as a strong foundation for downstream reasoning tasks.

\begin{figure*}
    \centering
    \includegraphics[width=\linewidth]{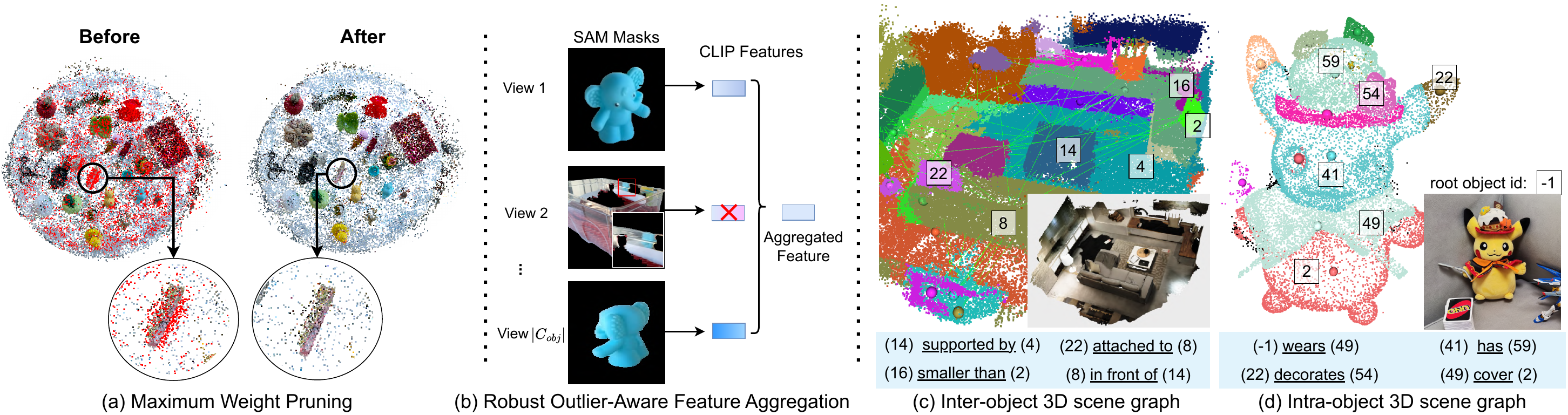}
    \caption{
    Illustration of proposed two improvement methods for hierarchical scene construction and two example scene graphs. 
    \textbf{(a):} 
    Low contribution Gaussian points (red) are removed to improve scene geometry.
    \textbf{(b):}
   Outlier features (e.g., due to occluded or inconsistent masks) are filtered before aggregation, producing a more coherent and consistent embedding for the target.
   \textbf{(c):} The spatial relationships are predicted by our GNN. 
   \textbf{(d):} The more semantic-enriched relationship lifted with LLM, the root object is marked as -1. }
    \label{fig:MCP_BA} \vspace{-5mm}
\end{figure*}

\noindent \textbf{Maximum Weight Pruning}. 
While THGS constructs its hierarchy directly from the raw Gaussian field, we observe that the unfiltered representation often contains numerous \textit{floaters}---Gaussians that contribute negligibly to the rendered views.  
These floaters, typically found near object boundaries or in occluded regions, introduce spurious pixel-aligned language features and lead to fragmented or noisy object-level clusters.  
Such artifacts become increasingly detrimental in multi-level hierarchical grouping, where boundary precision is crucial for maintaining semantic consistency across abstraction levels (see Fig.~\ref{fig:MCP_BA} (a)).  
To address this limitation, we introduce a \textit{maximum weight pruning} strategy that removes inconsistent or redundant Gaussians prior to hierarchical construction, thereby improving boundary integrity and the overall stability of the semantic hierarchy.

\vspace{4pt}
Building on Eq.~\ref{eq:max_contrib}, we compute the maximum contribution of each Gaussian across all camera views as:
\begin{equation}
    \omega_i^{\max}= \max_{c \in \mathcal{C}, p \in \mathcal{P}_c} \, w_{i,p}^{(c)},
\end{equation}
where $w_{i,p}^{(c)}$ denotes the contribution weight of Gaussian $i$ for pixel $p$ in camera view $c$. Here, $\mathcal{C}$ represents the set of training camera views, and $\mathcal{P}_c$ denotes the set of pixels in camera view $c$.
Gaussians whose maximum contribution weights falls below a small threshold $\tau_{contrib}$ are pruned from the scene:
\begin{equation}
    \mathcal{G}' = \{\, G_i \in \mathcal{G} \mid \omega_i^{\max} > \tau_{contrib}\,\}.
\end{equation}
This filtering step removes geometrically inconsistent or redundant Gaussians arising from occlusion or low opacity regions, while preserving the overall rendering quality.  
As demonstrated in our ablation study (Sec.~\ref{sec:eval:ablation}), the proposed pruning substantially improves geometric consistency and enhances the reliability of subsequent object-level clustering.

\noindent \textbf{Robust Outlier-Aware Feature Aggregation}.
Multi-view language registration often suffers from inconsistent CLIP features caused by erroneous SAM masks or extreme viewpoints. Directly averaging features across views, as done in THGS makes the object embedding sensitive to these outliers (see Fig.\ref{fig:MCP_BA} (b).
Given an object visible in $\mathcal{C}_{obj}$ views with corresponding CLIP features $\{\mathbf{f}_i\}_{i=1}^{|\mathcal{C}_{obj}|}$, we first measure the semantic consistency of each feature by computing its mean cosine similarity to all others:
\begin{equation}
s_i = \frac{1}{|\mathcal{C}_{obj}|-1} \sum_{j \ne i} \frac{\mathbf{f}_i \cdot \mathbf{f}_j}{\|\mathbf{f}_i\| \|\mathbf{f}_j\|}.
\end{equation}
We then perform Z-score normalization over the similarity scores:
\begin{equation}
z_i = \frac{s_i - \mu_s}{\sigma_s},
\end{equation}
where $\mu_s$ and $\sigma_s$ denote the mean and standard deviation of $\{s_i\}$. Features with $z_i < -\tau_{lang}$ are treated as low-similarity outliers and removed. Here, $\tau_{lang}$ is a Z-score threshold that controls the sensitivity of outlier detection. The final object-level embedding is obtained by averaging the remaining filtered features.
Our Robust Outlier-Aware Feature Aggregation (ROFA) suppresses multi-view inconsistencies and yields more reliable language-aligned object representations. As shown in our ablation study (Sec.~\ref{sec:eval:ablation}), this filtering leads to more stable and semantically consistent object embeddings.

\subsection{Relation Lifting and Prediction}
\label{sec:app:scene_graph}

Our hierarchical Gaussian scene enables explicit 3D scene graphs at different abstraction levels, as illustrated in Fig.~\ref{fig:MCP_BA} (c) and (d), a novel capability that brings structured, interpretable, and multi-level reasoning to Gaussian fields. 
At the top level $L$ (object level), we define an \textit{inter-object} graph capturing semantic relationship:
\begin{equation}
    \mathcal{H}_{\text{3D}}^{(L)} = (\mathcal{V}^{(L)}_{\text{obj}}, \mathcal{E}^{(L)}_{\text{inter}}, F_V^{(L)}, F_E^{(L)}),
\end{equation}
where $\mathcal{V}^{(L)}_{\text{obj}}$ and $\mathcal{E}^{(L)}_{\text{inter}}$ denote object nodes and their pairwise relations, with corresponding node and edge features $F_V^{(L)}$ and $F_E^{(L)}$.
Each node $v_k^{(L)}$ inherits its language-aligned feature $\mathbf{f}^{(L)}_k$, while each edge $(v_i,v_j)\!\in\!\mathcal{E}^{(L)}_{\text{inter}}$ carries a relation embedding $\mathbf{r}^{(L)}_{ij}$ encoded by Jina~\cite{sturua2024jina}.
At lower hierarchy levels $l<L$, we form \textit{intra-object} graphs to describe relationships among parts within the same object:
\begin{equation}
\begin{aligned}
    \mathcal{H}_{\text{3D,intra}}^{(l)} &= (\mathcal{V}^{(l)}_{\text{part}}, \mathcal{E}^{(l)}_{\text{intra}}, F_V^{(l)}, F_E^{(l)}), \\
    \mathcal{E}^{(l)}_{\text{intra}} &= \{(v_i,v_j)\,|\, v_i,v_j \in S^{(L)}_m \},
\end{aligned}
\end{equation}
where $S^{(L)}_m$ is the Gaussian cluster of object $m$.
The inter-object graph $\mathcal{H}_{\text{3D}}^{(L)}$ captures global scene relations, while the intra-object graphs $\mathcal{H}_{\text{3D,intra}}^{(l)}$ encode local part composition and affordance structure.
We next describe how both relation types can be either lifted using LLM-based annotation or predicted using pretrained GNN reasoning.

\noindent \textbf{3D Consistent Relation Lifting from LLM Annotation}.
To obtain open-vocabulary relational annotations, we follow the SoM–LLM paradigm also used in RelationField~\cite{koch2025relationfield}, but differ in how 2D object masks are acquired. 
In~\cite{koch2025relationfield}, per-frame instance masks is generated with SAM~\cite{kirillov2023segment}, which often suffer from view inconsistency and mixed semantic granularity.
In contrast, we leverage our hierarchical Gaussian scene and Eq.~\ref{eq:max_contrib} to render per-view Gaussian ID maps, trace them to their corresponding clusters, and thus obtain 
view-consistent cluster ID maps. 
%
Each cluster mask is then overlaid with numeric marks with the SoM prompting strategy~\cite{yang2023set}, enabling a MLLM such as GPT-4V to infer relationships between the marked object pairs.
For every pair of objects, the LLM outputs textual predicates $\langle s, p, o\rangle$ (subject, predicate, and object) as 2D relational annotations.
Because each mask corresponds to a 3D-consistent cluster ID, lifting these annotations into 3D is straightforward: we iterate over all annotated relations and assign them to the corresponding cluster pairs. 
After processing all frames, we pick the top-$k_p$ frequent predicates of an edge from all gathered predicates, encode the texts using Jina~\cite{sturua2024jina} and average to form the lifted relational embedding $\mathbf{f}_{ij}$. 

\noindent \textbf{Relation Prediction with GNN.} 
As explained in Sec.~\ref{sec:intro}, the lifted 3D scene graph has limited coverage of all plausible relationship in the scene. 
On the other hand, the latent space of open vocabulary spatial relationships is less diverse than the latent space describing the texture, shape, material and use of objects.
This motivates us to distill the open vocabulary relationship reasoning into a light weight pretrained GNN. 
We first construct a neighboring graph $\mathcal{H}'$ by connecting object nodes within a fixed distance threshold.
On top of this graph, we use the residual graph neural network $\mathcal{F}_{\theta}$ to predict relation embeddings:
\begin{equation}
    \hat{\mathbf{f}}_{ij} = \mathbf{f}_{ij}' + \mathcal{F}_{\theta}(\mathbf{f}_v^{\text{src}}, \mathbf{f}_v^{\text{dst}}, \mathbf{f}_{ij}'),
\end{equation}
where $\mathbf{f}_v^{\text{src}}$ and $\mathbf{f}_v^{\text{dst}}$ denote the language–geometry fused node embeddings of the source and destination objects (see Appendix~\ref{sec:appendix:gnn} for more details).
The network outputs open-vocabulary edge features $\hat{\mathbf{f}}_{ij} \!\in\! \mathbb{R}^{d_r}$ with the same embedding space as Jina-Embedding-V3, enabling direct cosine-similarity comparison with textual predicates.
We pretrain $\mathcal{F}_{\theta}$ on the 3RScan dataset~\cite{wald2019rio} following the setup of Open3DSG~\cite{koch2024open3dsg}, using a contrastive learning objective between predicted and ground-truth relation embeddings to align features across modalities.
Because the modality gap between language-lifted Gaussians and point-cloud–image features is small, the pretrained model generalizes well to our Gaussian domain and can be applied directly for inference without fine-tuning, as shown in Sec.~\ref{sec:eval:graph_querying}.

\subsection{Applications: Hierarchical and Relational Reasoning in 3D}
Our hierarchical scene representation and explicit 3D scene graph jointly enable structured reasoning within Gaussian fields.
The hierarchy supports \emph{compositional reasoning} by capturing how parts form objects (additional analysis of hierarchical queries at object and part levels is provided in Appendix \ref{sec:supp:hierarchical_eval}), while the scene graph facilitates \emph{relational reasoning} by modeling how objects interact.
To the best of our knowledge, we are the first to unify these two forms of reasoning within Gaussian fields, providing a causal understanding of the scene that links “what exists,” “how it is composed,” and “how it relates to each other.”
We demonstrate this capability through 
multi-hierarchy querying and triplet-based relational querying.

\noindent\textbf{Language-based multi-hierarchy querying.}
In practical situations involving natural language, people may describe either an entire object (\textit{``ramen''}) or one of its parts (\textit{``noodle''}), and may also refer to multiple similar instances simultaneously.
To handle these ambiguities, our goal is twofold: (1) automatically determine whether the query best matches a root-level object or a finer leaf-level cluster, and (2) identify all relevant matches when multiple clusters exhibit comparable similarity.
We design a tree-searching method to solve the querying problem analytically, following a simple intuition: a text referring to the sub-object part (leaf cluster) has higher similarity to the leaf language feature than the root.
We compute the cosine similarity of the CLIP embedding of a text query $\mathbf{t}$ with per-cluster features $\mathbf{f}^{(l)}_k$ across hierarchy levels.
Unlike the flat top-$k$ retrieval in~\cite{dai2025training,jiang2025votesplat}, we search over the nested cluster structure $\mathcal{S}^{(1 : L)}$ to adaptively match query granularity.
Starting from root-level candidates, the search descends to child clusters whenever they exhibit higher similarity, distinguishing whether the query refers to a whole object or one of its parts.
We further detect the largest drop in the similarity curve to automatically select multiple valid matches when present.
The resulting union of Gaussians forms an adaptive segmentation mask that unifies coarse object localization and fine part-level discovery within a single query, see Algm.~\ref{alg:query_in_tree} in Appendix.

\noindent \textbf{Language-Triplet querying on 3D Scene Graph.}
Beyond single-entity segmentation, we extend our querying framework to handle relationship-based queries of the form $\langle s, p, o\rangle$. 
Given such a query, our goal is to retrieve the 3D Gaussians corresponding to the subject entities that satisfy the specified relationship with the object.
To achieve this, we operate on the 3D scene graph constructed atop our multi-hierarchical scene representation. The search for potential subjects and objects follows the same multi-hierarchy querying procedure described above, ensuring that both fine-grained clusters and high-level object groups are considered as valid candidates.
After identifying candidate subject–object pairs, we evaluate their relationships by comparing the corresponding edge embeddings to the predicate embedding from the query. Each valid pair is then ranked using three complementary similarity measures: (1) subject–text alignment, (2) object–text alignment, and (3) predicate–relation alignment in the Jina embedding space. The Gaussians associated with the highest-ranked subject candidates are finally returned as the 3D regions most consistent with the queried relationship.

\section{Experiments}
\label{sec:experiments}
In this section, we evaluate our method from two aspects: first, 3D scene graphs including its prediction in Sec.~\ref{sec:eval:scene_graph_pred} and the extended application, relationship-guided 3D object segmentation using the scene graph in Sec.~\ref{sec:eval:graph_querying}, and second, the open vocabulary object segmentation and 3D semantic segmentation in Sec.~\ref{sec:eval:op_querying}. The best and second-best results of all experiments are marked in \textcolor{bestgreen}{green} and \textcolor{secondblue}{blue}, respectively.
We also provide ablation studies on our proposed different modules in Sec.~\ref{sec:eval:ablation}. Implementation details are given in Appendix~\ref{sec:apendix:imp_detail}.

\subsection{3D Scene Graph Prediction}
\label{sec:eval:scene_graph_pred}

We evaluate our method on the RIO10 subset of 3DSSG~\cite{wald2019rio} for 3D scene graph prediction.
This dataset provides semantic 3D scene graphs for pre-segmented 3D point clouds with 160 object classes and 27 relationship categories.
We compare against recent open-vocabulary 3D scene graph approaches, including Open3DSG~\cite{koch2024open3dsg}, which operates directly on pre-segmented point clouds, ConceptGraph~\cite{gu2024conceptgraphs}, which incrementally reconstructs scenes and scene graphs from RGB-D sequences, and RelationField~\cite{koch2025relationfield}.
We group them into scene-specific and scene-agnostic methods in Tab.~\ref{tab:scenegraph_evaluation_rio}, the same in Sec.~\ref{sec:eval:graph_querying}.
Quantitative results are reported with Recall@K for object, and relationship prediction following the protocols of~\cite{lu2016visual, yang2018graph}. 
Despite ground-truth annotation in~\cite{wald2019rio} exhibits uneven semantic granularity across scenes (e.g., carpets are merged with floors, while doors and door frames labeled separately) and sometime misaligned with our hierarchical representation,
our method with GNN still achieves better Recall on relationship prediction than others, with an improvement of 0.3 R@3 and 0.5 R@5 over RelationField, and exhibits only a 0.1 worse object Recall.
VLM-based method with 2D images as input performs poorly, due to the lack of spatial reasoning capability. 
Our method is also 4.7$\times$ faster and 7.6$\times$ more memory efficient than RelationField, see Tab.~\ref{tab:runtime_resource}.

\begin{table}[h]
\tabcolsep=1.75mm
\centering \vspace{-2mm}
\caption{\textbf{Results of 3D scene graph prediction on 3DSSG~\cite{wald2019rio}.} “Scene agnostic” denotes methods without per-scene training.}
\scalebox{0.99}{
\scriptsize
\begin{tabular}{@{}l|cc|cc|c@{}}
\toprule
\multirow{2}{*}{\textbf{Method}}  & \multicolumn{2}{c}{\textbf{Object}} & \multicolumn{2}{c
}{\textbf{Predicate}} & \multirow{2}{*}{\shortstack{\textbf{Scene}\\ \textbf{agno.}} }\\
   &  R@5 & R@10 &  R@3 &  R@5  & \\ 
\midrule

$\text{ConceptGraphs}$  \cite{gu2024conceptgraphs}    &0.37 & 0.46 & 0.74 & 0.79  & \XSolidBrush\\
RelationField~\cite{koch2025relationfield}   &  \best 0.69 & \best 0.80 & \sbest 0.76 & \sbest 0.82 &  \XSolidBrush \\
\rowcolor{gray!10}
Ours (VLM)  & \sbest 0.68 & \sbest 0.79 & 0.10 & 0.35 &\XSolidBrush\\
\midrule
Open3DSG \cite{koch2024open3dsg}                         & 0.56 & 0.61 & 0.58 & 0.65  & \checkmark\\
\rowcolor{gray!10}
Ours (pred.)  & \sbest 0.68 & \sbest 0.79 & \best 0.79 & \best 0.87 & \checkmark\\

\bottomrule
\end{tabular}
}

\label{tab:scenegraph_evaluation_rio} \vspace{-4mm}
\end{table}

\subsection{Evaluation on Open Vocabulary Instance Segmentation with Relationship Guidance}
\label{sec:eval:graph_querying}

We evaluate our method on \textbf{ScanNet++}~\cite{yeshwanth2023scannetpp} dataset with the benchmark provided by RelationField~\cite{koch2025relationfield}. 
The task aims to segment the \textit{subject} entity in 3D given an open-vocabulary relationship query represented as a triplet $\langle s, p, o\rangle$.
We disable both densification and our pruning as suggested in OpenGaussian~\cite{wu2024opengaussian}, to guarantee bijection between points in the ground truth point cloud and our Gaussian scene.
So that we can compute 3D mean-IoU between segmented object in Gaussian scene and the ground truth point cloud.

\begin{figure}[t]
 \centering \vspace{-1mm}
    \includegraphics[width=\linewidth]{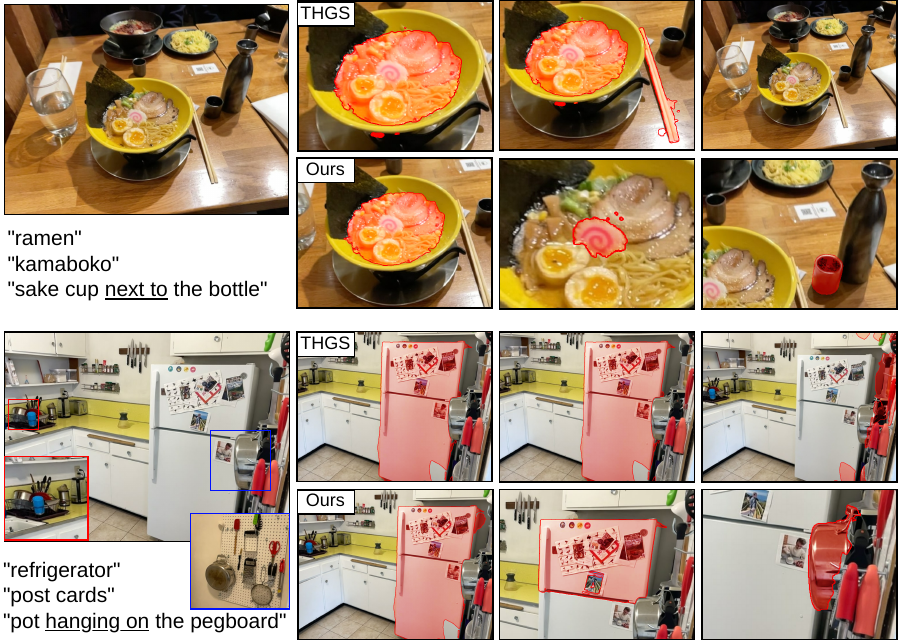}
        \caption{
        \textbf{Qualitative results of open vocabulary object segmentation.} 
        We show results on LERF dataset for segmentation mask on 2D view. 
        With multi-hierarchy querying search and 3D scene graph for relation guidance, our method shows strong improvement against THGS. 
        }
        \label{fig:querying} \vspace{-5mm}
\end{figure}

Among all language field distillation methods~\cite{engelmann2024opennerf, kerr2023lerf, qin2024langsplat, koch2025relationfield} we compared, only RelationField~\cite{koch2025relationfield} and our approach are using scene graph for relationship-based queries searching. 
Other methods are given a single query concatenated from the triplet text.
We use predicted 3D inter-object scene graph from our GNN, to investigate the cross-dataset generalization of our trained network.  
As shown in Tab.~\ref{tab:relation_seg}, our method achieves the highest performance among all approaches, surpassing both training-based and training-free baselines. 
In particular, our approach attains an mIoU of 0.56, outperforming the closest training-based competitor RelationField (0.53), despite requiring no additional training or LLM annotations. 
This improvement is attributed to our \textbf{multi-hierarchical scene representation} and \textbf{graph-based relational reasoning}, which together enable effective inference of spatial relationships purely from the 3D structure (see Fig.~\ref{fig:querying}). 
Removing the multi-hierarchy (w/o MH) degrades performance, underscoring the role of hierarchical search in resolving complex spatial relationships.
\begin{table}[htb]
\centering \vspace{-2mm}
\caption{
\textbf{Relationship-Guided 3D Instance Segmentation on ScanNet++~\cite{yeshwanth2023scannetpp}}.  
Our method achieves the highest mean IoU, even operating only on single hierarchy (w/o MH).
}
\scriptsize
\setlength{\tabcolsep}{3.5mm}
\begin{tabular}{l|c|c}
\toprule
\textbf{Method} & \textbf{mIoU} & \textbf{\textbf{Scene agno.}} \\
\midrule
Lerf~\cite{kerr2023lerf} & 0.25 & \XSolidBrush \\
OpenNeRF~\cite{engelmann2024opennerf} & 0.45 &  \XSolidBrush\\
LangSplat~\cite{qin2024langsplat} & 0.49 &  \XSolidBrush\\
RelationField~\cite{koch2025relationfield} & 0.53 &  \XSolidBrush\\
\midrule
THGS~\cite{dai2025training} & 0.29 & \checkmark \\
\rowcolor{gray!10}
Ours w/o MH & \sbest 0.54 & \checkmark \\
\rowcolor{gray!10}
Ours & \best 0.56 & \checkmark \\
\bottomrule
\end{tabular}
\label{tab:relation_seg} \vspace{-4mm}
\end{table}

\subsection{Open Vocabulary Object Querying}
\label{sec:eval:op_querying}

We evaluate our approach on two open-vocabulary segmentation benchmarks: LeRF-OVS~\cite{kerr2023lerf} for object segmentation and ScanNet~\cite{dai2017scannet} for semantic segmentation.
Tab.~\ref{tab:2D_lang_querying} and~\ref{tab:segmentation_results} report quantitative comparisons against recent methods, with training-free method specified (T-F).
Our method achieves state-of-the-art performance on both datasets, outperforming both training-based and training-free approaches.
On LeRF-OVS, we further analyze performance on hierarchical queries by separating object-level and part-level queries; detailed results are reported in Appendix \ref{sec:supp:hierarchical_eval}.
The most notable gains are observed in the \textit{Figurines} and \textit{Teatime} scenes, where the proposed feature aggregation and geometry refinement yield stronger semantic alignment under cluttered and occluded setups. We also provide qualitative results in Fig.~\ref{fig:querying}.
On ScanNet, our method achieves a slight improvement over 
This moderate gain is mainly attributed to the experiment setup, in which computing 3D-mIoU requires fixed Gaussian primitives number thus excluding our \textit{Maximum Weight Pruning} component. Detailed analysis using a revised evaluation protocol, including densified Gaussian reconstruction and nearest-point assignment, is provided in Appendix \ref{sec:supp:scanNet_additional_analysis}.

\begin{table}[htb]
\centering  \vspace{-2mm}
\caption{
\textbf{Open-vocabulary segmentation on LERF-OVS~\cite{kerr2023lerf}}.  
Our method achieves the highest mean IoU among both training and training-free approaches.
}
\scriptsize
\setlength{\tabcolsep}{2mm} 
\begin{tabular}{l | cccc c | c}
    \toprule
    \multirow{2}{*}{\textbf{Method}} & \multicolumn{4}{c}{\textbf{LERF-OVS mIoU (\%)}} & \multirow{2}{*}{\textbf{Mean}} & \multirow{2}{*}{\textbf{T-F}} \\
    \cmidrule(lr){2-5}
     & Fig. & Ramen & Teatime & Waldo & & \\
    \midrule
    LangSplatV2~\cite{li2025langsplatv2} & 56.4 & 51.4 & \sbest 72.2 & 59.1 & 59.9 & \XSolidBrush \\
    LAGA~\cite{cen2025tackling} & \sbest 64.1 & \best 55.6 & 70.9 & \best 65.6 & \sbest 64.0 & \XSolidBrush \\
    \midrule
    THGS~\cite{dai2025training} & 57.3 & 43.5 & 68.3 & 50.7 & 54.9 & \checkmark \\
    Occam’s~\cite{cheng2024occamslgssimpleapproach} & 58.6 & 51.0 & 70.2 & \sbest 65.3 & 61.3 &  \checkmark \\
    VALA~\cite{wang2025visibility} & 59.9 & \sbest 51.5 & 70.2 & 65.1 & 61.7 &  \checkmark \\
    \rowcolor{gray!10}
    \textbf{Ours} & \best 64.7 & 51.2 & \best 81.0 & 60.6 & \best 64.4 & \checkmark \\
    \bottomrule
\end{tabular}
\label{tab:2D_lang_querying}   \vspace{-3mm}
\end{table}

\begin{table}[htb] 
\centering  \vspace{-2mm}
\caption{
\textbf{Open-vocabulary semantic segmentation on ScanNet~\cite{dai2017scannet}}. 
Our method achieves the highest mean performance among both training and
training-free approaches on 15 and 10 classes subset.
}
\scriptsize
\setlength{\tabcolsep}{1.8mm} 
\begin{tabular}{l | c c c c c c | c}
\toprule
\multirow{2}{*}{\textbf{Methods}} & 
\multicolumn{2}{c}{\textbf{19 cls}} & 
\multicolumn{2}{c}{\textbf{15 cls}} & 
\multicolumn{2}{c}{\textbf{10 cls}} & 
\multirow{2}{*}{\textbf{T-F}} \\
\cmidrule(lr){2-3} \cmidrule(lr){4-5} \cmidrule(lr){6-7}
 & mIoU & mAcc & mIoU & mAcc & mIoU & mAcc & \\
\midrule
LangSplat~\cite{qin2024langsplat} & 3.78 & 9.11 & 5.35 & 13.20 & 8.40 & 22.06 & \XSolidBrush\\
OpenGau.~\cite{wu2024opengaussian} & 24.73 & 41.54 & 30.13 & 48.25 & 38.29 & 55.19 & \XSolidBrush\\
LAGA~\cite{cen2025tackling} & \sbest 32.50 & 49.10 & 35.50 & 53.50 & 42.60 & 63.20 & \XSolidBrush\\

\midrule

THGS~\cite{dai2025training} & \best 34.39 & \best 50.74 & \sbest 39.61 & \sbest 57.07 & \sbest 46.38 & 64.74 & \checkmark\\
Occam's~\cite{cheng2024occamslgssimpleapproach} & 31.93 & 48.93 & 34.25 & 53.71 & 45.16 & 64.39 & \checkmark\\
VALA~\cite{wang2025visibility} & 32.11 & \sbest 50.05 & 35.10 & 54.77 & 46.21 & \sbest 65.61 & \checkmark\\

\rowcolor{gray!10}
\textbf{Ours} & 32.35 & 44.52 & \best 40.04 & \best 60.59 & \best 47.17 & \best 66.08 & \checkmark \\
\bottomrule
\end{tabular}

\label{tab:segmentation_results} \vspace{-2mm}
\end{table}

\subsection{Ablation Study}
\label{sec:eval:ablation}
We evaluate the impact of our proposed components on open-vocabulary 2D segmentation (Tab.~\ref{tab:ablation_LeRF_full}). Maximum Weight Pruning yields the largest improvement, underscoring the importance of refining the scene geometry by removing low-contribution Gaussians that otherwise introduce structural noise. Robust Outlier-Aware Feature Aggregation provides additional gains by filtering inconsistent language features across views, improving the semantic coherence of the aggregated embeddings. Notably, its effect is most pronounced on the \textit{Figurines} and \textit{Ramen} scenes of LeRF-OVS, where dense object arrangements and frequent occlusions (and, in \textit{Ramen}, the presence of transparent surfaces) tend to corrupt the averaged semantic features. 

To further analyze the sensitivity of the outlier detection threshold $\tau_{\text{lang}}$, we report results for varying values in Tab.~\ref{tab:ablation_LeRF_BA}. The IoU initially increases with larger $\tau_{\text{lang}}$ as more consistent features are retained, but declines once the threshold becomes too permissive and introduces noise.

A runtime and memory consumption analysis is given in Tab.~\ref{tab:runtime_resource}, in which we decompose the resource usage of our method in three stages, showing that our method can inject structure semantic representation to a Gaussian scene with less than 25\% memory overhead. More ablation studies are given in the appendix about the graph neural network design, the pruning's effect on rendering quality, and the effect of our pruning and language lifting method on the final scene graph prediction.

\begin{table}[htb]
\centering
\caption{
\textbf{Ablation} on LeRF-OVS dataset for open-vocabulary segmentation. $M.$ is Maximum Weight Pruning and $R.$ is Robust Outlier-Aware Feature Aggregation .
} 
\scalebox{0.75}
    {
        \begin{tabular}{ l |c c c c|c }
        \toprule
        Variant & Figurines & Ramen & Teatime & Kitchen & Mean \\ 
        \midrule
        Base & 52.05 & 47.19 & 76.77 & 47.5 & 55.88 \\
        Base+$M.$ & 59.16 & 47.41  & 80.98 & 60.59 & 62.04\\
        Base+$M.$+$R.$(Full) &  64.69 & 51.15  & 80.98 &	60.6 &	64.36 \\
        \bottomrule
        \end{tabular}
    }
\label{tab:ablation_LeRF_full} \vspace{-4mm}
\end{table}

\begin{table}[htb]
\centering
\caption{
\textbf{Ablation} on different values of $\tau_{lang}$ on LeRF-OVS.
}
\scalebox{0.8}
    {
        \begin{tabular}{ c |c c c c|c }
        \toprule
        $\tau_{lang}$ & Figurines & Ramen & Teatime & Waldo Kitchen & Mean \\ 
        \midrule
        1 & 61.85 & 43.3 & 76.02  & 63.37 & 60.78 \\
        2 & 63.39 & 51.07 & 73.8  & 60.62 & 62.22\\
        3 &  64.69	&51.15 & 80.98 &	60.60 &	64.36 \\
        4 & 63.3 &  47.4 & 80.98 & 60.60 & 63.07 \\
        
        \bottomrule

        \end{tabular}
    }

\label{tab:ablation_LeRF_BA} \vspace{-4mm}

\end{table}

\begin{table}[htb]
    \centering
    \caption{\textbf{Runtime and resource comparison} against RelationField~\cite{koch2025relationfield} on 3DSSG scenes. Our method is 4.7$\times$ faster and 7.6$\times$ more memory efficient and a maximum of 7.5GB of GPU memory.}
    \scalebox{0.75}{
    \begin{tabular}{l|c c c | c} 
    \toprule
        \multirow{2}{*}{\textbf{Method}} & \multicolumn{4}{c}{\textbf{Time (min) / Storage (MB) / GPU Memory (GB)}} \\ 
        & Scene Rec. & Language Distill. & Scene Graph & Overall \\ \midrule
        RF~\cite{koch2025relationfield} & \multicolumn{3}{c|}{60/500/32 (inseparable)} & 60/500/32\\ 
        Ours & 11/52.2/3.2 & 1.5/10.6/7.5 & 0.1/2.2/3.0 & 12.6/65/7.5 \\  \bottomrule
    \end{tabular}}
    \label{tab:runtime_resource} \vspace{-3mm}
\end{table}

\section{Conclusion}
\label{sec:conclusion}

We introduced ReLaGS, the first framework to unify multi-hierarchical 3D Gaussian fields and open-vocabulary 3D scene graphs within a single language-grounded representation.
Through  \emph{Maximum Weight Pruning} and \emph{Robust Outlier-Aware Feature Aggregation}, ReLaGS improves geometric fidelity, multi-view language registration, and the overall robustness of the scene representation.
Building upon this unified scene representation, our explicit scene graph formulation enables scalable, training-free relational reasoning at far lower computational and memory costs than prior methods.
ReLaGS achieves state-of-the-art performance across open-vocabulary segmentation, 3D scene graph prediction, and relationship-guided object segmentation, advancing open-vocabulary 3D understanding and paving the way for causal and compositional reasoning in complex real-world scenes.
We hope our framework enables the community to efficiently generate large-scale, semantically enriched 3D radiance fields at low computational and resource cost.

\clearpage
\section*{Acknowledgements}

This work has been partially funded by the EU projects
dAIEDGE (Grant Agreement Number: 101120726) and LUMINOUS (Grant Agreement Number: 101135724).

{
    \small
    \bibliographystyle{ieeenat_fullname}
    \bibliography{main}
}

\clearpage
\setcounter{page}{1}
\maketitlesupplementary

\noindent\textbf{Appendix Overview.}
This appendix provides additional technical details, implementation notes, ablations, and extended results supporting the main paper. 
We first present \textbf{the architecture of our graph neural network} (Sec.~\ref{sec:appendix:gnn}), including geometric feature construction, node/edge encoders, the edge-aware transformer, and the contrastive training objective. 
Sec.~\ref{sec:apendix:imp_detail} describes \textbf{implementation details} for Gaussian reconstruction, feature extraction, pruning, language aggregation, and LLM-based annotation. 
\textbf{Additional ablation studies} (Sec.~\ref{sec:supp:more_ablation}) evaluate the impact of pruning, feature aggregation, and GNN components. 
Sec.~\ref{sec:supp:scene_graph_prediction} details \textbf{the evaluation protocol for 3D scene graph prediction}, including object–cluster matching and closed-set readout. 
We then provide further explanation of our multi-hierarchy querying algorithm (Sec.~\ref{sec:supp:query}), followed by extended qualitative and quantitative results across all benchmarks (Sec.~\ref{sec:supp:more_results}). 
Finally, Sec.~\ref{sec:supp:datasets} and Sec.~\ref{sec:supp:metrics} summarize datasets and evaluation metrics used throughout our experiments.

\section{Graph Neural Network Architecture}
\label{sec:appendix:gnn}

To infer open-vocabulary relationships between object pairs, we train a geometry–language fused graph transformer that operates on the 3D scene graph derived from our hierarchical Gaussian representation. Each object (or part) is treated as a node, and each potential relationship forms a directed edge. The goal of the network is to output a 512-dimensional relation embedding for each edge, compatible with the Jina-Embedding-V3 predicate space.

\noindent \textbf{Node and Edge Geometric Feature Construction.}
Each object node is represented by a 19-dimensional geometric descriptor computed from its oriented bounding box (OBB).  
Given an object's 3D points, we estimate its center $\mathbf{c}_i$, PCA rotation matrix $\mathbf{R}_i \in \mathbb{R}^{3\times 3}$, OBB half-extents $\mathbf{d}_i$, and mass center $\mathbf{m}_i$.  
These components are concatenated into:
\begin{equation}
    \mathbf{g}_i = 
\big[
\mathbf{c}_i \,\Vert\, \mathbf{m}_i \,\Vert\,
\mathbf{R}_i(:,1) \,\Vert\, \mathbf{R}_i(:,2) \,\Vert\, \mathbf{R}_i(:,3) \,\Vert\, 
\mathbf{d}_i
\big] \in \mathbb{R}^{19},
\end{equation}
that captures the object’s location, orientation, scale, and height.
For each directed edge $(i,j)$, we compute a corresponding relational feature $\mathbf{g}_{ij}\in \mathbb{R}^{19}$ using the OBBs of both nodes.  
This includes relative displacement $\mathbf{c}_j - \mathbf{c}_i$, Euclidean distance, normalized direction vector, height difference, PCA-axis alignment, and simple OBB overlap/support cues.  
These geometric features  allow the GNN to distinguish relations such as \emph{left}, \emph{behind}, \emph{supported by}, and \emph{inside} purely from spatial structure.

\noindent \textbf{Node and Edge Encoders.}
Each node fuses its CLIP feature $\mathbf{f}_i$ and geometric descriptor $\mathbf{g}_i$ through a single MLP, and each directed edge $(i,j)$ is encoded by concatenating the two node features with its geometric edge descriptor $\mathbf{g}_{ij}$:
\begin{equation}
\begin{aligned}
\mathbf{f}_v^{(i)} &= \mathrm{MLP}_{\text{node}}\!\left(\mathbf{f}_i \,\Vert\, \mathbf{g}_i\right), \\[2pt]
\mathbf{f}_{ij} &= \mathrm{MLP}_{\text{edge}}\!\left(\mathbf{f}_v^{(i)} \,\Vert\, \mathbf{f}_v^{(j)} \,\Vert\, \mathbf{g}_{ij}\right),
\end{aligned}
\label{eq:gnn_encoder}
\end{equation}
where $\mathbf{f}_v^{(i)}, \mathbf{f}_{ij} \in \mathbb{R}^{512}$.
This provides unified node and edge embeddings that jointly capture semantic and geometric information before graph message passing.

\noindent \textbf{Edge-Aware Graph Transformer.}
For relational reasoning, we adopt a lightweight edge-aware graph transformer inspired by SGFormer~\cite{lv2024sgformer}.
Unlike the original classification-oriented design, our variant predicts \emph{open-vocabulary relation embeddings} instead of closed-set predicate logits.
The transformer operates jointly on node and edge embeddings.
At each layer, node updates are computed using \emph{edge-aware attention} in which the edge feature $\mathbf{f}_{ij}$ modulates the attention from node $i$ to node $j$.
This allows the model to incorporate geometric cues directly into the attention weights, improving discrimination of fine-grained spatial relations.

In parallel, edge features are iteratively refined.
Each edge embedding receives a gated residual update conditioned on both endpoint nodes and its previous feature, enabling it to accumulate higher-order relational context across layers.
Overall, this transformer block follows the same structure as SGFormer—multi-head edge-aware attention, node aggregation, and edge refinement—but adapted to produce continuous 512D embeddings aligned with the Jina predicate space rather than discrete predicate labels.

\noindent \textbf{Training Strategy.}
We supervise predicted relation embeddings using a simple multi-positive contrastive objective.  
For an edge $(i,j)$ with predicted embedding $\hat{\mathbf{f}}_{ij}$, ground-truth predicate embeddings $\mathcal{P}$, and a set of sampled negatives $\mathcal{N}$, the loss is:
\begin{equation}
\mathcal{L}_{\mathrm{ctr}}
=
-
\log
\frac{
\sum_{p\in\mathcal{P}}
\exp\!\left(\hat{\mathbf{f}}_{ij}\cdot \mathbf{f}_p / \tau\right)
}{
\sum_{p\in\mathcal{P}}
\exp\!\left(\hat{\mathbf{f}}_{ij}\cdot \mathbf{f}_p / \tau\right)
+
\sum_{n\in\mathcal{N}}
\exp\!\left(\hat{\mathbf{f}}_{ij}\cdot \mathbf{f}_n / \tau\right)
}.
\end{equation}

This loss naturally supports multiple correct predicates per edge and encourages separation from unrelated or antonymic relations. The resulting embedding space captures fine-grained open-vocabulary spatial semantics for 3D scene graph prediction.

\noindent \textbf{Generalization to Gaussian Scenes.}
Although the network is trained exclusively on 3RScan using point-cloud geometry and RGB-derived CLIP features, it generalizes effectively to Gaussian scenes because:
(i) geometric descriptors are representation-independent,
(ii) language embeddings come from CLIP and Jina, which are modality-agnostic,
(iii) relational reasoning primarily depends on spatial configuration rather than texture,
and (iv) the InfoNCE loss aligns all relation types into a shared embedding space.
As a result, the pretrained GNN can be applied directly to our Gaussian scene graphs without any fine-tuning and produces robust relation predictions.

\section{Implementation Details}
\label{sec:apendix:imp_detail}
The Gaussian scenes used in all experiments are trained with 2DGS~\cite{Huang2DGS2024} following THGS~\cite{dai2025training}.
For semantic feature extraction, we use SAM ViT-H for 2D segmentation guidance in all experiments, while two types of CLIP are used for different tasks: 
(1) For open-vocabulary object querying, we follow LangSplat~\cite{qin2024langsplat} using OpenCLIP ViT-B/16 ($\mathbf{f}_k \in \mathbb{R}^{512}$) as the vision-language encoder.
(2) For the task of 3D scene graph prediction and relationship-guided 3D instance segmentation, we follow RelationField to use the vision-language feature from CLIP/ OpenSeg~\cite{ghiasi2022scaling} ($\mathbf{f}_k \in \mathbb{R}^{768}$) and use jina-embedding-v3~\cite{sturua2024jina} to encode the language feature of relationships ($\mathbf{f}_{ij} \in \mathbb{R}^{512}$).
We introduce two hyperparameters in our hierarchical reconstruction and language lifting pipeline. 
For Maximum Weight Pruning, we use a contribution threshold $\tau_{contrib}=5\times10^{-4}$, to remove geometrically inconsistent Gaussians without affecting the rendering quality of the scene. 
For Robust Outlier-Aware Feature Aggregation, we set $\tau_{lang}=3$ to remove outlier language features. 
For lifting 3D scene graph from LLM-annotations, we use $K_p=3$ to pick the top 3 frequent relationships per edge for encoding the open-vocabulary edge embedding. 
The relationship annotation from 2D images is extracted using GPT-4o~\cite{achiam2023gpt}. 
We train our graph neural network on 3DSSG dataset, with all testing sequences in RIO10 subset excluded to prevent data leakage, with a batch size of 4.
We begin with a 20-epoch warm-up phase using only the cosine similarity loss, followed by 60 epochs of training with contrastive loss.
The network contains only one transformer layer and is trained for 80 epochs with learning rate $1\times 10^{-4}$.
We run all experiments of our method using a single NVIDIA RTX 3090 GPU. 

\section{Hierarchical Query Evaluation on LeRF}
\label{sec:supp:hierarchical_eval}

A key design goal of our framework is to support hierarchical semantic querying, enabling retrieval at different semantic granularities such as object-level entities (e.g., ramen) and object parts (e.g., noodles). While the LeRF dataset contains queries spanning both levels, the dataset does not explicitly distinguish between object-level and part-level queries. To analyze this capability more precisely, we manually categorize the LERF queries across scenes into object and part queries.

Using this categorization, we evaluate retrieval performance separately for the two groups. The results are summarized in Tab. \ref{tab:hierarchical_reasoning}. Our method shows clear improvements over the baseline across both categories, with particularly strong gains on part-level queries. Specifically, our approach improves part-level mIoU by more than \textbf{15\%}, highlighting the effectiveness of our hierarchical scene representation in capturing fine-grained semantic structure.

\begin{table}[htb] 
\centering
\caption{\textbf{Hierarchical query evaluation on LERF.} We manually categorize LERF queries into object-level and part-level queries and report mIoU for each category. Our method improves performance across both groups, with particularly large gains on part-level queries, demonstrating the ability of the proposed representation to support fine-grained hierarchical language queries.}
\vspace{1mm}
    \scalebox{0.9}{
    \begin{tabular}{l | cc}
        \hline
        Method & Object mIoU & Part mIoU \\
        \hline
        THGS &  63.2 &  45 \\
        Ours & \best 65.3 & \best \textbf{60.3} \\
        \hline
    \end{tabular}}
    \vspace{-5mm}
    \label{tab:hierarchical_reasoning}
\end{table}

\section{Additional Analysis on ScanNet}
\label{sec:supp:scanNet_additional_analysis}
Training Gaussian scenes on ScanNet using a fixed number of primitives leads to suboptimal reconstruction quality, as previously observed in Dr.~Splat~\cite{jun2025dr} (see App.~C, Fig.~S2).
Applying Maximum Weight Pruning (MWP) with $\tau_{\text{contrib}} = 0$ removes a substantial fraction of Gaussian primitives, which can further degrade reconstruction quality.
To address this, we enable densification during reconstruction (denoted by ) and, at evaluation, assign each Gaussian the ground-truth label of its nearest point. Our protocol is conceptually similar to that of Dr.~Splat, which uses the Mahalanobis distance; we instead use the standard L2 distance, which simplifies computation while preserving the evaluation behavior.
Using this protocol for both THGS and our method, we observe consistent improvements of our approach over THGS across all class subsets (Tab.~\ref{tab:scanNet_new_protocol}).

\begin{table}[htb]
    \centering
    \caption{\textbf{ScanNet Results with Revised Evaluation Protocol.} Quantitative results on ScanNet using densified Gaussian reconstruction and nearest-point assignment (L2). THGS* refers to the THGS method re-run by us on all scenes under this protocol. Our method consistently outperforms THGS* across all class subsets.}
    \scalebox{0.9}{
    \begin{tabular}{l | cc cc cc}
        \hline
        \multirow{2}{*}{Method} 
        & \multicolumn{2}{c}{19 Classes} 
        & \multicolumn{2}{c}{15 Classes} 
        & \multicolumn{2}{c}{10 Classes} \\
        \cline{2-7}
        & mIoU & mAcc & mIoU & mAcc & mIoU & mAcc \\
        \hline
        THGS & 39.33 & 54.23 & 43.61 & 61.55 & 52.52 & 70.50 \\
        Ours & \best 41.26 & \best 57.29 & \best 45.24 &\best 62.75 &\best 54.04 &\best 71.48 \\
        \hline
    \end{tabular}}
    \vspace{-3mm}
    \vspace{-4mm}
    \label{tab:scanNet_new_protocol}
\end{table}

\section{More Ablation Studies}
\label{sec:supp:more_ablation}

\paragraph{MWP and ROFA on Scene Graph Prediction.}
We provide the ablation study of the proposed Max Weight Pruning (MWP) and Robust Outlier-Aware Feature Aggregation (ROFA) methods on the task of 3D scene graph prediction on 3DSSG~\cite{wald2019rio} dataset in Tab.~\ref{tab:appendix_ablation_scene_graph}.
The graph neural network for 3D scene graph edge prediction relies on the object geometry and semantic feature as input, therefore, accurate the geometry segmentation and language feature lifting significantly improve the object and edge classification results. 
Our pruning method brings a large performance improvement on 3DSSG dataset, where the original quality of the reconstructed Gaussian scenes is relatively low, due to the low-resolution and motion-blur of the dataset's RGB video sequences.

\begin{table}[htb]
    \centering
        \caption{\textbf{Ablation} on 3DSSG dataset for 3D scene graph prediction. M. is Maximum Weight Pruning and R. is Robust Outlier-Aware Feature Aggregation .}
    \scalebox{0.9}{
    \begin{tabular}{@{}l|cc|cc@{}}
    \toprule
    \multirow{2}{*}{\textbf{Variants}}  & \multicolumn{2}{c}{\textbf{Object}} & \multicolumn{2}{c
    }{\textbf{Predicate}} \\
       &  R@5 & R@10 &  R@3 &  R@5  \\ 
    \midrule
      Base  & 0.57 & 0.69 & 0.75 & 0.85  \\ 
      Base+$M.$  & 0.65 & 0.75 & 0.77 & 0.87  \\
      Base+$M.$+$R.$  & 0.68 & 0.79 & 0.79 & 0.87 \\
    
    \bottomrule
    \end{tabular}
    }
    \label{tab:appendix_ablation_scene_graph}
\end{table}

\paragraph{Graph Neural Network on Scene Graph Prediction.}

We conduct an ablation study on different design variants of our graph neural network and evaluated them on 3DSSG dataset in Tab~\ref{tab:appendix_ablation_scene_graph_gnn}.
As the results show, adding more transformer layers does not improve performance and instead makes the network significantly heavier. This indicates that local information in the scene graph is more important than long-range context for reasoning about relational edges.
Furthermore, we also test the different design choices of providing the language feature of nodes to the edge encoder (as in Eq.\ref{eq:gnn_encoder}) than initializing edge feature from zeros.
As the result shows, initializing edge feature from zero performance less good than giving the edge encoder semantic hints from its connected nodes.

\begin{table}[htb]
    \centering
        \caption{\textbf{Ablation} on 3DSSG dataset for 3D scene graph prediction. L is the number of transformer layer, and $0  \mapsto e$ means using zero vector instead of using node feature pair to initialize edge feature $n  \mapsto e$.}
    \scalebox{0.9}{
    \begin{tabular}{@{}l|cc|cc@{}}
    \toprule
    \multirow{2}{*}{\textbf{Variants}}  & \multicolumn{2}{c}{\textbf{Object}} & \multicolumn{2}{c
    }{\textbf{Predicate}} \\
       &  R@5 & R@10 &  R@3 &  R@5  \\ 
    \midrule
      L=1, $n  \mapsto e$ & 0.68 & 0.79 & 0.79 & 0.87  \\ 
      L=2, $n  \mapsto e$ & 0.68 & 0.79 & 0.77 & 0.84  \\
      L=1,  $0  \mapsto e$ & 0.68 & 0.79 & 0.75 & 0.88 \\
    
    \bottomrule
    \end{tabular}
    }
    \label{tab:appendix_ablation_scene_graph_gnn}
\end{table}

\paragraph{Pruning on Rendering Quality.}

We conduct an ablation study to examine how maximum weight pruning affects both the rendering quality of the 3D scene and the downstream open-vocabulary segmentation performance. As shown in Tab. \ref{tab:pruning_threshold_ablation}, applying a light pruning threshold ($\tau_{contrib}=5\times10^{-4}$) significantly improves mIoU from 57.84 to 64.36, while leaving all rendering metrics unchanged. This behavior reflects the role of pruning in removing Gaussians that contribute negligibly to the rendering quality but introduce geometric noise. Eliminating these low-weight Gaussians leads to a cleaner and more accurate scene geometry, which directly benefits the multi-hierarchical clustering step used before lifting language features into the Gaussian clusters.

As the pruning threshold increases, the geometric simplification becomes more aggressive. Moderate thresholds ($5\times10^{-3}$ and $5\times10^{-2}$) begin to slightly perturb the geometry, leading to a gradual decrease in mIoU while still having almost no effect on rendering quality. However, at high thresholds (e.g. $5\times10^{-1}$), pruning removes a substantial portion of the scene structure, which severely degrades both geometry and photometric fidelity. This collapse propagates to the semantic stage, resulting in a steep drop in segmentation performance. These results demonstrate that lightweight pruning is essential to enhance geometric consistency and, in turn, improve the quality of multi-hierarchical grouping and language feature lifting, while overly aggressive pruning compromises the underlying scene representation.

\begin{table}[htb]
\centering
\caption{\textbf{Ablation Study} on weight pruning.}
\scalebox{0.9}{
\begin{tabular}{ c |c|ccc }
\toprule
$\tau_{contrib}$ & mIoU $\uparrow$ & SSIM $\uparrow$ & PSNR $\uparrow$ & LPIPS $\downarrow$ \\ \hline
- & 57.84 & 0.8478 & 23.576 & 0.2492 \\
$5\times10^{-4}$ & 64.36 & 0.8478 & 23.576 & 0.2492 \\
$5\times10^{-3}$ & 59.27 & 0.8478 & 23.576 & 0.2492 \\
$5\times10^{-2}$ & 57.53 & 0.8462 & 23.526 & 0.2516 \\
$5\times10^{-1}$ & 39.18 & 0.4967 & 10.905 & 0.5185 \\
\bottomrule
\end{tabular}
}
\label{tab:pruning_threshold_ablation}
\end{table}

\section{3D Scene Graph Prediction on 3DSSG}
\label{sec:supp:scene_graph_prediction}
\noindent \textbf{Ground Truth and Prediction Alignment.}
When evaluating on 3DSSG, we face two main challenges.
First, our scene graph is constructed on a Gaussian field rather than a fixed, pre-segmented point cloud.
In RelationField~\cite{koch2025relationfield}, semantic features can be directly queried at any 3D location within a NeRF, whereas in our Gaussian-based representation, we establish a nearest-neighbor mapping between ground-truth 3D points and their corresponding Gaussians.
Second, the ground-truth segmentation in 3DSSG exhibits uneven semantic granularity across scenes. For example, carpets are often merged with floors, while doors and door frames may be labeled separately.
To ensure fair comparison, we search across all hierarchical levels of our Gaussian clusters and select, for each ground-truth object, the cluster whose 3D oriented bounding box achieves the highest IoU overlap.
After establishing the object–cluster correspondences, we apply our trained graph neural network to predict open-vocabulary relationship features between these clusters.

\noindent \textbf{Closed-set Label Readout.}
Although our node features $\mathbf{f}_v^{(i)}$ and relation features $\hat{\mathbf{f}}_{ij}$ live in open-vocabulary embedding spaces (CLIP for objects, Jina for relations), inference is performed over a fixed closed-set of dataset labels.
For objects, each class name $c_k$ from the closed vocabulary $\mathcal{C}$ is encoded once with the CLIP text encoder to get $\mathbf{t}_k$.  
Classification is obtained by a cosine-similarity projection,
$\mathrm{score}(i,k)=\langle \mathbf{f}_v^{(i)},\mathbf{t}_k\rangle$,
followed by a softmax over $k$.
Similarly, each predicate name $r_m$ from the closed predicate set $\mathcal{R}$ is embedded with the Jina text encoder into $\mathbf{q}_m$.  
Relation prediction for edge $\mathcal{E}_{(i,j)}$ is obtained by 
$\mathrm{score}(ij,m)=\langle \hat{\mathbf{f}}_{ij},\mathbf{q}_m\rangle$,
again softmax-normalized across $m$.
In both cases,
we take the closed-set ground-truth label whose embedding is most similar to the node/edge feature (in the corresponding CLIP/JINA space) and assign it as the top-ranked label. This provides a clean way to map open-vocabulary features back to dataset-specific closed labels.

\vspace{3pt}
\noindent \textbf{Evaluation under Dense Graph Construction.}
Different from prior works on 3DSSG~\cite{koch2024open3dsg, koch2025relationfield}, which typically evaluate triplet recall using fixed thresholds such as Recall@50 or Recall@100, our method constructs a denser spatial scene graph.
Specifically, we connect object pairs within a fixed distance threshold (5\,m), motivated by the observation that many spatial relationships can potentially exist between nearby objects.
However, the human-annotated relations in 3DSSG are relatively sparse, meaning that only a small subset of valid spatial relations is labeled.
Under a dense graph construction, the number of candidate triplets grows rapidly with the number of objects and predicate classes, which makes fixed Recall@$K$ metrics less informative.
Therefore, in addition to standard Recall@$K$, we also report \textbf{Triplet Recall@1\% and 5\%} of the ranked predictions.
This percentage-based metric normalizes for the size of the candidate relation space and provides a more stable measure of relational reasoning performance under dense graph settings.
Using this evaluation, our method achieves a Triplet Recall of \textbf{0.86 at 1\%} and \textbf{0.94 at 5\%}, demonstrating strong relational reasoning performance even under a dense graph formulation.

\section{Further Detail about Multi-hierarchy Querying}
\label{sec:supp:query}
Alg.~\ref{alg:query_in_tree} provides the procedural formulation of the multi-hierarchy querying mechanism described in the main paper.
The algorithm implements our key idea that the correct semantic granularity of a text query—whether it refers to a whole object or to one of its finer parts—can be inferred by comparing similarity scores across hierarchy levels.
Starting from root-level clusters, the procedure iteratively descends the tree only when child clusters exhibit higher CLIP similarity to the query, thereby adapting the search depth to the level implied by language.
To handle cases where several sibling clusters are equally relevant, the algorithm further analyzes similarity drops along the ranked candidates and selects all clusters above the largest drop.
The final segmentation mask is obtained by aggregating all Gaussians belonging to the retained clusters, matching the unified object–part reasoning discussed in the main paper.

\begin{algorithm}[thb]
\algsetup{linenosize=\tiny}
\SetAlFnt{\small}
\SetAlCapNameFnt{\small}
\caption{Querying in Hierarchical Scene}
\SetKwInOut{Input}{Input}
\SetKwInOut{Output}{Output}

\Input{
    Text query $q$; 
    ~Cluster features $\{\mathbf{f}^{(l)}_k\}_{l=1}^{L}$; \\
    ~Multi-level labels $\{\mathcal{S}^{(l)}\}$; 
    ~$K$ of Top-$k$.
}
\Output{
    Binary mask $\mathbf{M} \in \{0,1\}^{N}$ for Gaussians.
}

\BlankLine
\textbf{1.} Encode text query: $\mathbf{t} \leftarrow \mathrm{VLMEncode}(q)$;\\
\textbf{2.} Get similarity at all levels: $\rho^{(l)} = \mathrm{cos}(\mathbf{t}, \mathbf{f}^{(l)}_k)$;\\
\textbf{3.} Select top-$K$ root-level candidates $(l{=}L)$ at $\rho^{(L)}$;\\
\textbf{4.} For each root candidate $S^{(L)}_r$: \\
\Indp
    \textbf{(a)} Retrieve its child clusters $\{S^{(L-1)}_c\}$;\\
    \textbf{(b)} If $\max_c \rho^{(L-1)}_c > \rho^{(L)}_r$, \\
    \Indp descend one level and repeat (4). \Indm \\
    \textbf{(c)} Otherwise, keep $S^{(L)}_r$ as the matched cluster.\\
\Indm
\textbf{5.} Filter out clusters smaller than $1\%$ of parent size;\\
\textbf{6.} Detect largest score drop $\Delta \rho_{\max}$ and retain clusters above it;\\
\textbf{7.} Aggregate Gaussians belonging to the selected clusters: 
$\mathbf{M}[G_i]=1$ if $G_i \in S^{(l)}_{\text{selected}}$.

\BlankLine
\textbf{Return:} hierarchical segmentation mask $\mathbf{M}$.
\label{alg:query_in_tree}
\end{algorithm}

\section{More Results}
\label{sec:supp:more_results}
\noindent \textbf{Open Vocabulary Object Querying on 3D-OVS.}
We provide more quantitative result for open-vocabulary object segmentation on five scenes of the 3D-OVS dataset~\cite{liu2023weakly} in Tab.~\ref{tab:2D_lang_querying_3dovs}.
Following Occam'sLGS~\cite{cheng2024occamslgssimpleapproach}, we evaluate our method on the corrected annotation of 3D-OVS dataset on the "room" sequence.
3D-OVS dataset mainly contains image sequences of a few objects from close-up viewpoints without much occlusion, in which sharp and clear segmentation of object boundaries plays a critical role for IoU.
Although our method based on heuristic clustering performs suboptimal in this set-up and sometimes meets difficulty of providing sharp boundaries, we still achieve a huge improvement compared to THGS and a comparable performance with state-of-the-art methods.

\begin{table}[htb]
\centering  \vspace{-2mm}
\caption{
\textbf{Open-vocabulary segmentation on 3D-OVS~\cite{liu2023weakly}}.  
Methods with $*$ are evaluated on testing-data that has an annotation error in the 3D-OVS testset corrected. $\dagger$ means we run the open-source code under our environment.
}
\scriptsize
\setlength{\tabcolsep}{2mm} 
\begin{tabular}{l | cccc c c}
    \toprule
    \multirow{2}{*}{\textbf{Method}} & \multicolumn{5}{c}{\textbf{3D-OVS mIoU (\%)}} & \multirow{2}{*}{\textbf{Mean}}   \\
    \cmidrule(lr){2-6}
     & Bed & Bench & Lawn & Room & Sofa &  \\
    \midrule
    LangSplatV2~\cite{li2025langsplatv2} & 93.0 & 94.9 & 96.1 & 92.3 & 96.6 & 94.6 \\
    LAGA~\cite{cen2025tackling} & 96.8 & 92.8 & 97.0 & 93.0 & 96.9 & 95.3 \\
    \midrule
    Occam’s~\cite{cheng2024occamslgssimpleapproach} & 96.8 & 94.8 & 97.0 & 96.5* & 88.8 & 95.0 \\
    THGS~\cite{dai2025training}$\dagger$ & 68.9 & 67.6 & 73.9 & 68.4* & 54.6 & 66.7 \\
    \rowcolor{gray!10}
    \textbf{Ours} & 95.4 & 94.4 & 91.2 & 94.8* & 86.9 & 92.5 \\
    \bottomrule
\end{tabular}
\label{tab:2D_lang_querying_3dovs}   \vspace{-3mm}
\end{table}

\begin{figure*}
    \centering
    \includegraphics[width=\linewidth]{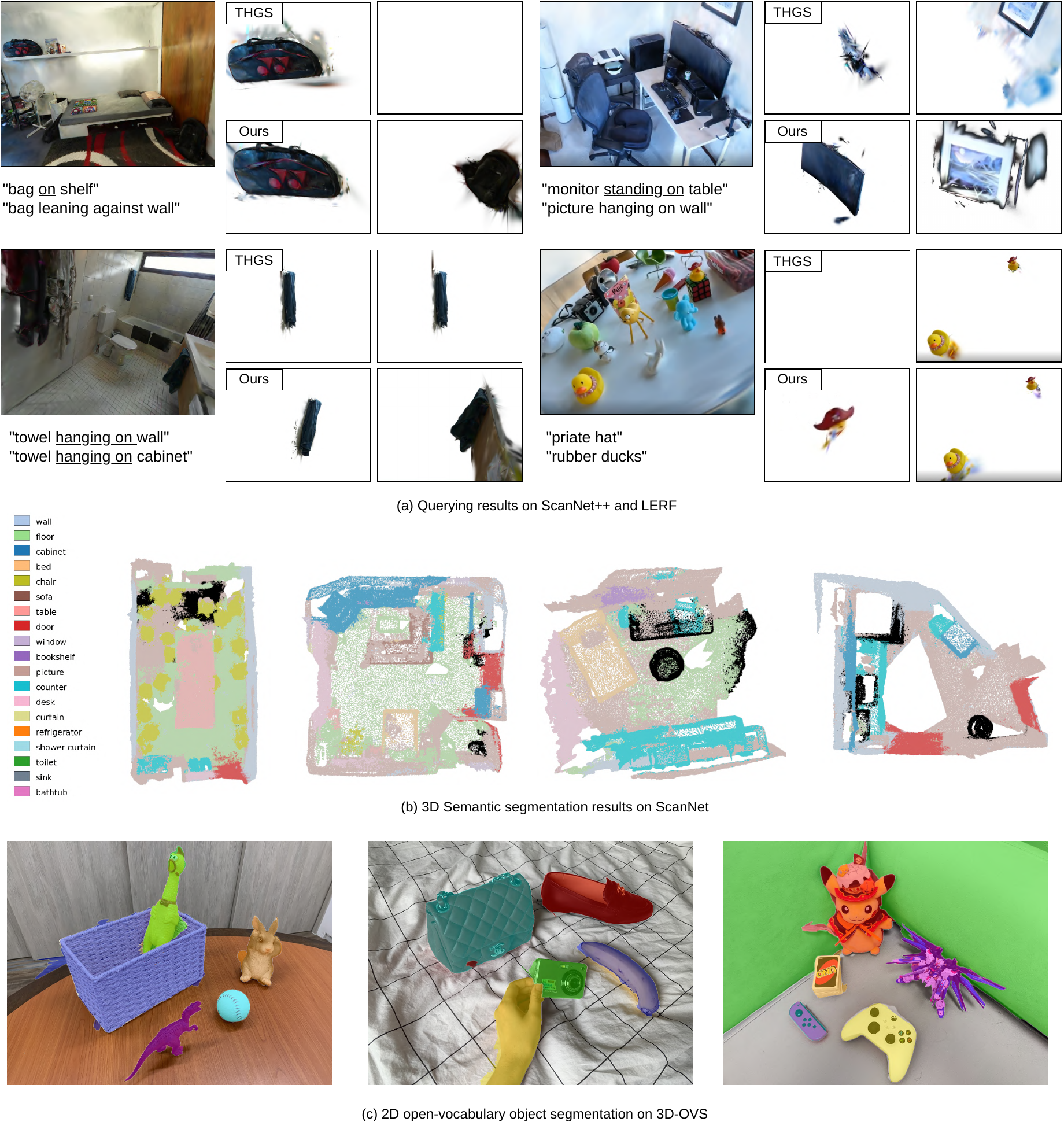}
    \caption{\textbf{Qualitative results} on LERF, ScanNet++, ScanNet and 3D-OVS.}
    \label{fig:supp_querying}
\end{figure*}

\begin{figure*}
    \centering
    \includegraphics[width=\linewidth]{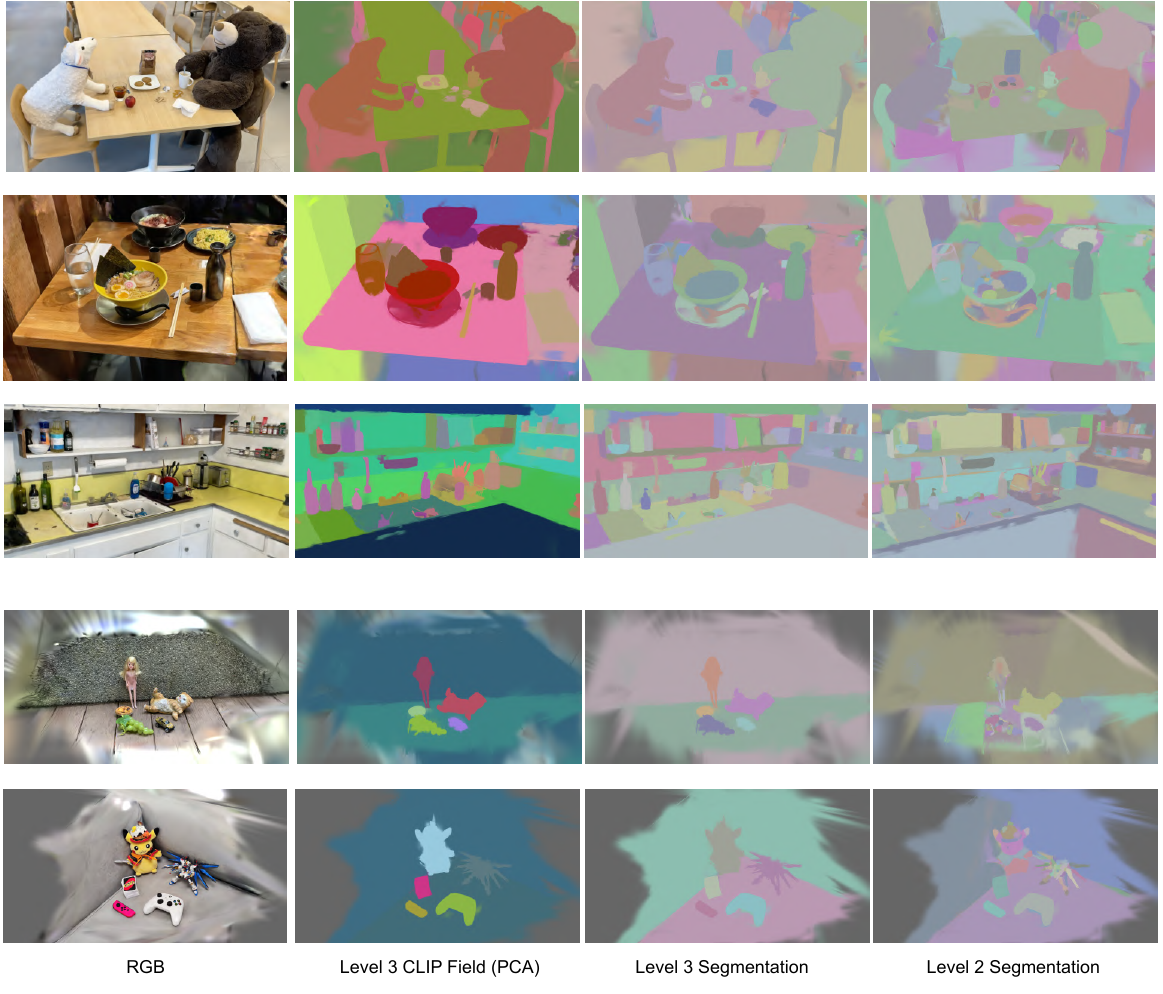}
    \caption{Visualization of multi-hierarchy scene reconstruction on LERF and 3D-OVS datasets.}
    \label{fig:supp_pca}
\end{figure*}

\begin{figure*}
    \centering
    \includegraphics[width=\linewidth]{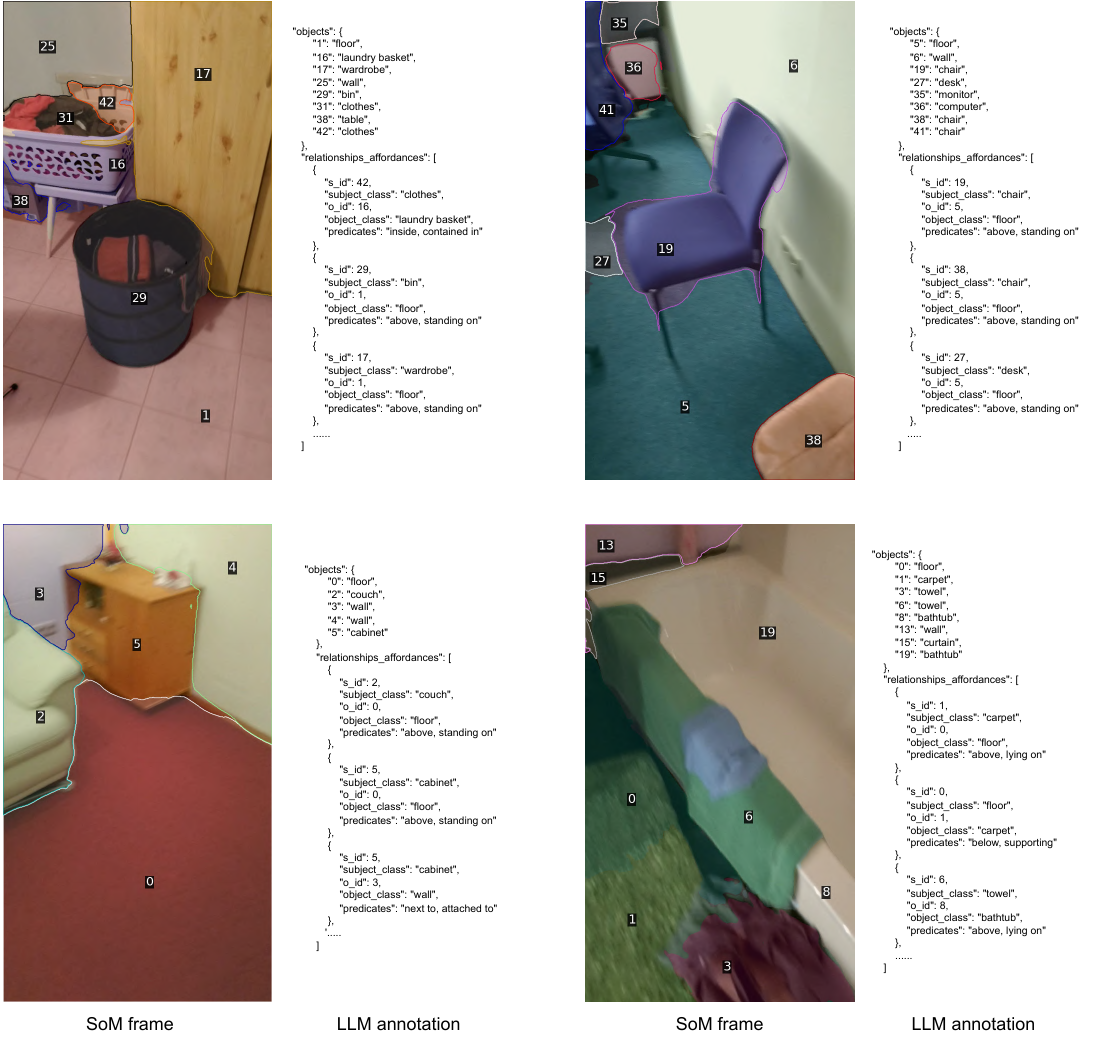}
    \caption{Examples of our SoM+LLM scene graph annotation on 2D images from 3DSSG dataset.}
    \label{fig:sup_som_anno}
\end{figure*}

We further provide more qualitative results on LERF, 3D-OVS, ScanNet and ScanNet++ dataset in Fig.~\ref{fig:supp_querying}, on the task of querying single object, object parts and also relation guided object querying. 
Visualization of the multi-hierarchy scene segmentation and PCA colored language field is given in Fig~\ref{fig:supp_pca}, as well as some examples of the SoM-LLM annotation with our lifting method in Fig.~\ref{fig:sup_som_anno}.

\section{Datasets}
\label{sec:supp:datasets}

\noindent\textbf{LERF-OVS.}
LERF \cite{kerr2023lerf} introduces language-embedded radiance fields for 2D open-vocabulary localization.
LERF-OVS adds object-level masks for quantitative evaluation via 2D mIoU.
It mainly tests fine-grained semantic alignment and part–object disambiguation in cluttered scenes.

\noindent\textbf{3D-OVS.}
3D-OVS \cite{liu2023weakly} provides RGB-D reconstructions with 3D ground-truth instance labels for
open-vocabulary segmentation.
Unlike LERF-OVS, it evaluates volumetric 3D semantic consistency.
We follow the corrected annotation for the ``room'' scene as in Occam’s LGS.

\noindent\textbf{ScanNet.}
ScanNet \cite{dai2017scannet} contains 1513 indoor RGB-D scenes with dense 3D semantic labels and is a standard benchmark for
3D semantic segmentation.
We follow prior work and report 3D mIoU on the 19/15/10-class subsets.

\noindent\textbf{ScanNet++.}
ScanNet++ \cite{yeshwanth2023scannetpp} refines ScanNet with higher geometric fidelity, improved trajectories,
and more accurate reconstructions.
It provides 3D instance annotations and serves as the benchmark for relationship-guided instance segmentation with annotations provided by RelationField\cite{koch2025relationfield}.

\noindent\textbf{3DSSG.}
3DSSG \cite{wald2019rio} includes 3D semantic scene graphs with 160 object categories and 27 relationship types
annotated on pre-segmented point clouds.
It is the standard dataset for 3D scene graph prediction.
We use the RIO10 subset and follow Open3DSG and RelationField for evaluating object and predicate recall,
and for training our lightweight GNN.

\section{Evaluation Metrics}
\label{sec:supp:metrics}

\noindent \textbf{Metrics for Open-Vocabulary Querying.}
For LERF-OVS and 3D-OVS, we evaluate open-vocabulary localization using the standard 2D mIoU protocol from LERF~\cite{kerr2023lerf} and OVS~\cite{wu2024opengaussian}.
Given a query $q$, each model renders a semantic heatmap from the viewpoint of the ground-truth mask.
We compute the per-pixel similarity between the CLIP embedding of $q$ and the rendered language features, threshold it into a binary prediction $P_q$, and measure:
\begin{equation}
\text{mIoU}(q) = \frac{|P_q \cap G_q|}{|P_q \cup G_q|},
\end{equation}
where $G_q$ is the ground-truth 2D region.
This 2D mIoU measures how well the language-aligned 3D features project to 2D and localize open-vocabulary concepts at the pixel level.

\vspace{3pt}
\noindent
For 3D benchmarks such as ScanNet~\cite{dai2017scannet} and ScanNet++~\cite{yeshwanth2023scannetpp}, we instead evaluate semantic correctness directly in 3D.
Each Gaussian (or reconstructed 3D point) is assigned a semantic label, and predictions are compared with ground-truth point-level annotations.
The 3D mIoU is computed using the intersection-over-union between predicted and ground-truth point sets:
\begin{equation}
\text{mIoU}_{3D} = \frac{|P \cap G|}{|P \cup G|},
\end{equation}
where $P$ and $G$ denote the sets of predicted and ground-truth points for a given semantic category.
Unlike 2D mIoU, which evaluates pixel-level consistency in single views, 3D mIoU measures volumetric semantic accuracy across the full reconstructed scene.
For ScanNet++, predictions are aligned to the ground-truth mesh to compensate for scale and sampling differences before evaluation.

\noindent \textbf{Metrics for 3D Scene Graph Prediction.}
We evaluate relational reasoning on the 3DSSG dataset~\cite{wald2019rio} following the standard scene graph metrics used in prior work~\cite{yang2018graph, lu2016visual, koch2024open3dsg, koch2025relationfield}.
A scene graph is defined by nodes (objects), edges (relationships), and their semantic labels.
Evaluation is performed using three complementary recall-based metrics: \textbf{object recall} and \textbf{predicate recall}.
All metrics report \textbf{Recall@K}, i.e., whether the correct ground-truth label appears in the top-$K$ predictions.

\noindent \textbf{Object Recall@K.}
For each ground-truth object, we compute the cosine similarity between its textual class name and the predicted object embedding.
The model outputs a ranked list of object class candidates, and an object is counted as correctly recognized if the ground-truth class appears among the top-$K$ predictions.
This evaluates the quality of open-vocabulary language registration and the discriminability of object-level embeddings.

\noindent \textbf{Predicate Recall@K.}
For each annotated relation edge $(s, p, o)$, the model predicts a ranked list of predicate labels using cosine similarity between the predicted relation embedding and the textual embeddings of all predicates.
Predicate Recall@K measures the fraction of ground-truth predicates appearing in the top-$K$ predictions.
Because many spatial predicates (e.g., \textit{next to}, \textit{in front of}, \textit{on top of}) may describe similar geometric configurations, this metric captures the model’s ability to recognize fine-grained and open-vocabulary.

\noindent \textbf{Triplet Recall@p\%.}
Triplet recall evaluates whether the full relational triplet $(s,p,o)$—including the subject, predicate, and object—is correctly predicted.
For each candidate object pair in the scene, the model produces scores for all predicate classes, forming a ranked list of candidate triplets.
All triplets are globally ranked by their confidence scores, and Triplet Recall@p\% measures the fraction of ground-truth triplets that appear within the top-$p\%$ of this ranked prediction set.
Reporting recall at a percentage rather than a fixed $K$ better reflects performance in dense graphs, where the number of candidate relations grows with the number of objects and predicate classes.

\section{Discussion on Dynamic Scenes and 4D Extensions}
\label{sec:supp:discussion_dynamic_scenes}

While our method focuses on static 3D scenes, the proposed language registration framework can be naturally extended to dynamic environments. In particular, adapting the approach to 4D scene representations would require incorporating temporal consistency into the clustering and relational reasoning mechanisms used in our pipeline.

A key component of our method is the construction of a KNN graph over Gaussians, which is subsequently used for hierarchical clustering and relational reasoning. In dynamic scenes, this graph must account not only for spatial proximity but also for temporal consistency across frames. The precise adaptation depends on the type of dynamic Gaussian representation used.

In approaches where scene dynamics are modeled through deformation networks applied to a fixed set of Gaussians, the number of Gaussians remains constant over time. In this setting, extending our method would primarily require redefining Gaussian proximity across time. Specifically, Gaussians belonging to the same object should maintain spatial consistency throughout the time intervals in which they are visible. Under this assumption, the KNN graph can be constructed using a temporal aggregation of spatial distances, allowing the hierarchical clustering procedure to remain largely unchanged. Importantly, our language registration components, including MWP and ROFA, can be applied without modification.

A second class of dynamic representations models scenes using 4D Gaussians, where primitives may appear or disappear over time. In this case, the graph construction must explicitly incorporate the temporal dimension, as spatial proximity alone is insufficient to determine relationships between primitives. Both the KNN graph and the subsequent hierarchical clustering would therefore need to operate in a joint spatio-temporal space, where edge weights reflect proximity across both spatial and temporal dimensions.

Finally, dynamic scenes introduce time-varying relationships between objects, which cannot be fully captured by a static scene graph. Extending our relational reasoning framework would therefore require the use of a dynamic scene graph that evolves over time, enabling the model to represent changing object interactions and relations.

Overall, these considerations suggest that the proposed framework provides a promising foundation for language-grounded reasoning in dynamic 4D scenes, with the primary extensions involving temporally-aware graph construction and clustering while leaving the core language registration mechanisms largely unchanged.

\end{document}